\documentclass[journal]{IEEEtran}

\ifCLASSINFOpdf
   \usepackage[pdftex]{graphicx}
\fi
\usepackage{subcaption}
\usepackage{lipsum}
\usepackage{booktabs}
\usepackage{float}
\usepackage{color}
\usepackage{amsmath}
\usepackage{amssymb}
\usepackage{algorithmic}
\usepackage{array}
\usepackage{cite}
\usepackage[colorlinks=true,
            linkcolor=blue,
            urlcolor=blue, 
            citecolor=blue]{hyperref}
\usepackage{url}
\usepackage{ntheorem}
\theoremseparator{:}
\newtheorem{theorem}{\textsl{Theorem}}
\theorembodyfont{\normalfont}

\theoremseparator{ }
\newtheorem{definition}{\textsl{Definition}}
\theorembodyfont{\upshape}

\theoremseparator{:}
\newtheorem{remark}{\textsl{Remark}}
\theorembodyfont{\upshape}

\theoremseparator{ }
\newtheorem{problem}{\textsl{Problem}}
\theorembodyfont{\upshape}

\theoremseparator{:}
\newtheorem{lemma}{\textsl{Lemma}}
\theorembodyfont{\upshape}

\begin{document}

\title{Dual-Stage Safe Herding Framework for Adversarial Attacker in Dynamic Environment}
\author{{\bf Wenqing Wang$^{1}$, Ye Zhang$^{2}$, Haoyu Li$^{3}$, Jingyu Wang$^{4}$}
}

\maketitle

\begin{abstract}
Recent advances in robotics have enabled the widespread deployment of autonomous robotic systems in complex operational environments, presenting both unprecedented opportunities and significant security problems. Traditional shepherding approaches based on fixed formations are often ineffective or risky in urban and obstacle-rich scenarios, especially when facing adversarial agents with unknown and adaptive behaviors. This paper addresses this challenge as an extended herding problem, where defensive robotic systems must safely guide adversarial agents with unknown strategies away from protected areas and into predetermined safe regions, while maintaining collision-free navigation in dynamic environments. We propose a hierarchical hybrid framework based on reach-avoid game theory and local motion planning, incorporating a virtual containment boundary and event-triggered pursuit mechanisms to enable scalable and robust multi-agent coordination. Simulation results demonstrate that the proposed approach achieves safe and efficient guidance of adversarial agents to designated regions.
\end{abstract}

\begin{IEEEkeywords}
Reach-Avoid game, target defense, multi-agent system, cooperative strategy
\end{IEEEkeywords}

\IEEEpeerreviewmaketitle

\section{Introduction}
\subsection{Motivation}

\IEEEPARstart{R}{ecently} advancements in robotics technology have led to the widespread deployment of autonomous unmanned aerial vehicles (UAVs) in complex operational environments, demonstrating substantial potential across diverse applications including search-and-rescue localization, surveillance mapping, and cooperative transportation.
However, while offering civilian benefits, these autonomous systems present significant tactical threats through military reconnaissance of sensitive areas and catastrophic infrastructure damage via explosive payloads targeting critical facilities such as airports. Conventional counter-drone strategies predominantly employ two approaches: hard-kill mechanisms involving physical interception through collision-based neutralization (e.g., interceptor UAVs or projectiles), and soft-kill techniques that disrupt communication links via signal jamming. Nevertheless, both methodologies exhibit critical limitations in urban low-altitude scenarios. Hard-kill implementations risk collateral damage to civilian structures and populations while incurring substantial operational costs. Moreover, the operational independence of autonomous UAVs from real-time ground control communications significantly reduces soft-kill effectiveness. This dual limitation necessitates the development of innovative alternative solutions for urban autonomous UAV countermeasures.\par
Autonomous adversarial agents (hereafter termed attackers) often exhibit intrinsic risk-averse characteristics to maintain survivability during mission execution. These agents implement evasion strategies, such as artificial potential field navigation, to actively avoid defensive entities (defenders) and environmental obstacles. This bidirectional action-response mechanism constitutes the fundamental interaction paradigm through which defenders achieve indirect motion control and positional guidance of attackers toward designated containment zones. Compared to hard-kill and soft-kill countermeasures, this methodology presents a safer and more cost-effective solution for autonomous UAV neutralization, offering significant theoretical and practical advantages.\par
Our research focuses on the challenge of safely catching and migrating an attacker with unknown evasion strategies. The core objective involves designing coordinated defensive strategies that simultaneously: 1)expel attackers from protected zones to predefined safe regions; 2)maintain collision-free navigation in obstacle-dense environments.This dual requirement necessitates novel formulations in multi-defender collaborative control under partial observability constraints.
\subsection{Related Works}
\textbf{Shepherding Problem}: Previous studies have established biologically inspired interaction models for adversarial agents, leveraging collective biological phenomena such as flocking \cite{strombomSolvingShepherdingProblem2014} and herding mechanisms \cite{paranjapeRoboticHerdingFlock2018}. These frameworks reformulate non-cooperative indirect guidance challenges as shepherding problems, yielding systematic solution methodologies. Those methods typically deploy defender agents in predefined encircling formations \cite{zhangCollectingFlockMultiple2022} to steer adversaries toward containment zones. Pierson and Schwager \cite{piersonControllingNoncooperativeHerds2018} pioneered potential field-based methods that simulate nonlinear repulsive forces, employing arc-shaped formations for directional control while addressing nonholonomic-to-holonomic kinematic mapping via offset controllers. Vásquez \cite{vasquezAdversarialScenariosHerding2023} extended similar formation strategies, quantitatively analyzing defender attrition impacts on mission success rates. Concurrently, Varava et al. \cite{varavaHerdingCagingTopological2017} proposed a "caged herding" approach using high-potential virtual barriers coordinated through RTT motion planning.
However, a critical limitation of these studies is their reliance on prior knowledge of adversary interaction patterns (including force dynamics and effective ranges). Practical scenarios frequently involve time-varying unknown parameters, exemplified by adaptive obstacle-avoidance radii \cite{songSpeedDensityPlanning2024} - smaller radii indicating aggressive tactics and larger radii corresponding to conservative strategies. Insufficient defender density relative to adversary aggressiveness may permit boundary breaches due to inadequate repulsive forces \cite{chipadeMultiSwarmHerdingProtecting2020}.
Addressing strategy uncertainty, Panagou et al. \cite{chipadeMultiagentPlanningControl2021,chipadeAerialSwarmDefense2023,zhangHerdingAdversarialSwarm2021} developed the "String Net" herding framework \cite{chipadeHerdingAdversarialSwarm2019}, postulating impenetrable boundaries through physical net structures. Their methodology combines mixed-integer programming with heuristic approaches for multi-swarm task allocation and intra-defender collision avoidance. Subsequent 3D environment validations \cite{zhangHerdingAdversarialSwarm2021} demonstrated operational feasibility. Nevertheless, two fundamental challenges remain: 1) Complex unmodeled dynamics arising from net deformation during motion, and 2) Practical implementation barriers for physical net structures in real-world deployments.\par
\textbf{Game-Theoretic Approaches}: Game-theoretic methodologies have emerged as promising frameworks for regional defense challenges, particularly through differential game theory. Originating from pursuit-evasion (PE) problem formulations in two-player non-cooperative conflicts \cite{zhangGameDronesMultiUAV2023,fangCooperativePursuitMultiPursuer2022,garciaDesignAnalysisStateFeedback2018,ramanaPursuitStrategyCapture2017,nardiGameTheoreticRobotic2018}, this paradigm derives Nash equilibrium strategies via Hamilton-Jacobi-Isaacs (HJI) equation solutions. While sharing adversarial dynamics with PE problems, non-cooperative indirect herding introduces heightened complexity through strict trajectory controllability requirements - unlike PE scenarios where terminal capture positions remain undetermined.
Recent advancements demonstrate diverse implementations: Zhang et al. \cite{zhangGameDronesMultiUAV2023} developed reinforcement learning-based pursuit strategies in obstacle-rich multi-agent PE games; Fang and Ramana \cite{fangCooperativePursuitMultiPursuer2022,ramanaPursuitStrategyCapture2017} established sufficient capture conditions through cooperative encirclement tactics against superior-speed adversaries; Nardi et al. \cite{nardiGameTheoreticRobotic2018} devised distributed control protocols solving multi-zone shepherding problems via open-source Nash equilibrium solvers. The reach-avoid (RA) game variant \cite{margellosHamiltonJacobiFormulation2011} extends PE frameworks by incorporating defensive zones, enabling methodological transfer to indirect herding challenges.
Notable innovations include Garcia's reachability analysis \cite{garciaCooperativeTargetProtection2021} circumventing HJI computations through Voronoi-based cooperative strategies \cite{zhouCooperativePursuitVoronoi2016}, and Yan's convex-domain RA game decomposition \cite{yanGuardingSubspaceHighDimensional2022,yanMatchingbasedCaptureStrategies2022,yanTaskAssignmentMultiplayer2020,yanReachAvoidGamesTwo2019} partitioning game spaces into offense/defense-dominant regions via singular barrier surfaces. Zhao et al. \cite{zhaoCollaborativeConstrainedTargetReaching2024} introduced virtual pipeline strategies decoupling path defense and blockade subgames, though their reliance on static obstacle assumptions and global environmental knowledge limits applicability in dynamic scenarios. 

\subsection{Overview of the Proposed Approach}
Through the above discussion, it can be observed that while significant progress has been made in shepherding problems and pursuit-evasion games, the non-cooperative indirect herding problem remains underexplored, and existing methods still face unresolved technical challenges. First, the non-cooperative indirect herding problem involves multi-agent game scenarios, requiring real-time strategy computation. Specifically, defensive agents must rapidly react to limited information about adversarial attackers to maintain defensive superiority. Second, motion planning for multi-agent systems in complex obstacle environments must be addressed to avoid collisions and safely navigate agents to desired regions. To tackle these challenges, this paper proposes a hierarchical hybrid strategy framework based on reach-avoid games and local motion planning to solve the non-cooperative indirect guidance problem.\par
\begin{figure}[h]
\centering
\includegraphics[width=0.85\linewidth]{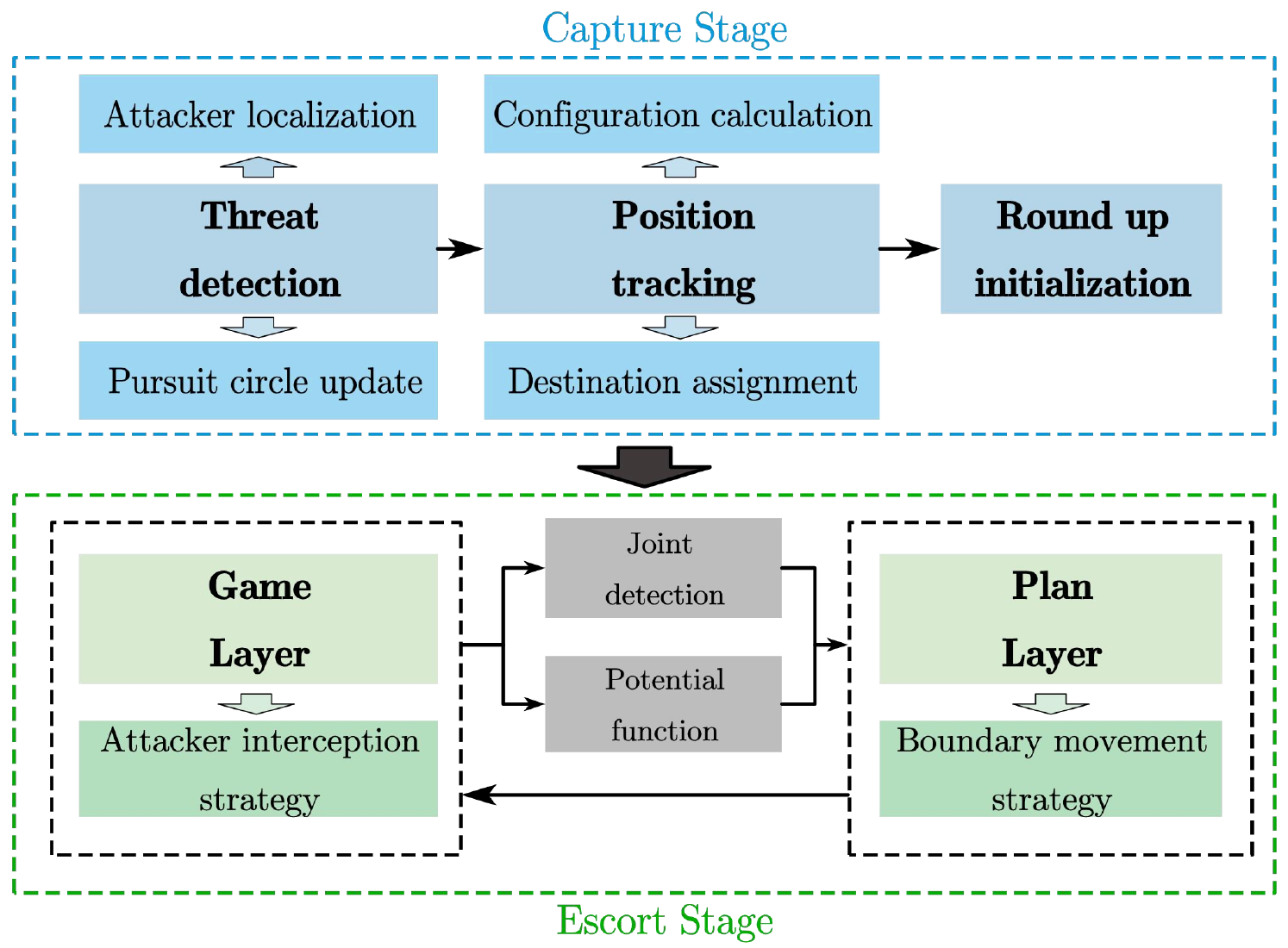}
\caption{Framework of indirect guidance approach.}
\label{Framework}
\end{figure}
As illustrated in Fig \ref{Framework}, our framework divides the area defense process into two stages: the capture stage and the escort stage. In the capture stage, defenders dispersed across different locations are required to form a predefined distribution configuration upon detecting attackers. To achieve this, we introduce an event-triggered pursuit circle method that converts the continuous motion trajectories of attackers into discrete sequences of circle centers, which can reduce the communication frequency between defenders and base stations. The desired coordinates of surrounding points are then calculated and assigned to defenders as inputs for their position-tracking controllers. It is important to note that a strategically rational configuration setup is critical for guaranteeing a successful game outcome, and the specific calculation method will be elaborated in subsequent sections. In the escort stage, defenders must respond instantaneously to attacker's behaviors to prevent escape and guide them safely toward the target area. To address the complexity of this dual-task scenario, we propose a resilient escort method that decouples the problem into a Game Layer and a Plan Layer. The plan layer employs a virtual beacon movement strategy to coordinate the collective movement direction of defenders, while the game layer integrates an attacker interception strategy to enable individual defenders to counteract escape attempts dynamically.

\subsection{Contributions}
In this article, we present an area defense method capable of defending against an attacker with unknown strategies in complex environments. Compared to previous studies, the key contributions of this work are as follows:
\begin{itemize}
    \item[$\bullet$] We investigate the understudied multi-agent non-cooperative indirect guidance problem in complex continuous environment including dynamic obstacles, and propose a scalable hierarchical hybrid strategy framework for cooperative capture and escort. By integrating physical defense UAVs and virtual containment boundary, this framework decomposes the complex non-cooperative indirect guidance problem into multiple tractable one-to-one reach-avoid (RA) subgames and local obstacle avoidance problems, which can be addressed independently;
    \item[$\bullet$] We partition the game space and derive analytical expressions for the barrier, rigorously analyze the regional reachability of attackers, and establish a clear situational judgment criterion alongside defenders' control objectives;
    \item[$\bullet$] We systematically examine the impact of defenders' control errors on game outcomes, construct relative error variables with bounded domain constraints, and develop a feasible computational method for initial distribution configuration parameters;
    \item[$\bullet$] Under this framework, a victory-guaranteed feedback countermeasure strategy is designed for physical defense UAVs based on relative error analysis and prescribed performance control. Unlike \cite{zhaoCollaborativeConstrainedTargetReaching2024}, we proposed a continuous control form for defenders, which eliminates the buffeting problem near the boundary, and there is no need for elaborately designed initial constraint parameters either. Simultaneously, we devise motion planning strategy for virtual beacons using joint detection and artificial potential field approach, which relies solely on local obstacle information. These strategies are analytical and computationally lightweight, enabling distributed deployment and real-time computation.
\end{itemize}

\section{Problem Formulation}
\subsection{Scenario description}

\begin{figure}
\centering
\includegraphics[width=\linewidth]{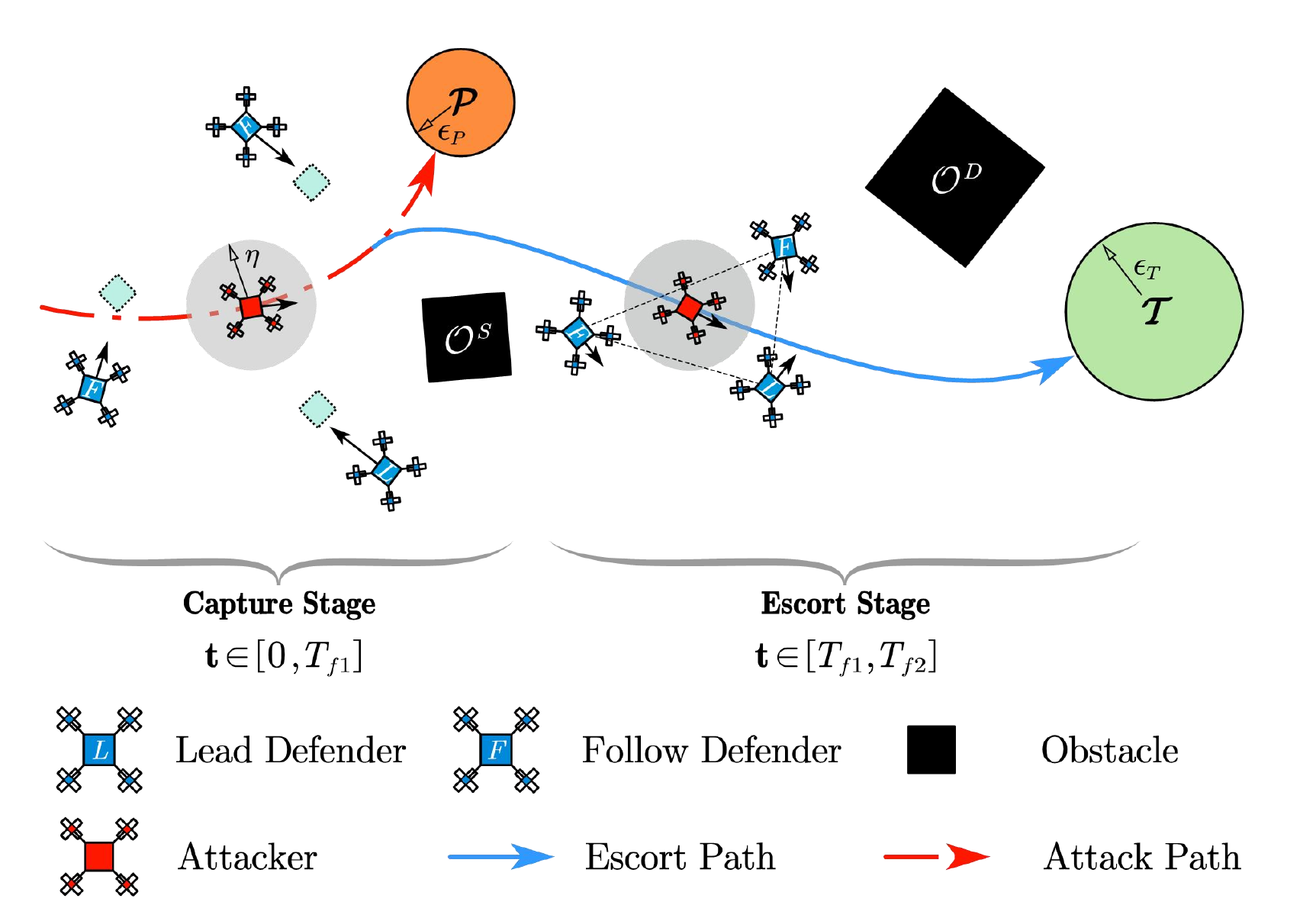}
\caption{Scenario Description}
\label{Scenario Description}
\end{figure}

We consider the non-cooperative indirect guidance problem in a complex two-dimensional euclidean space $\Omega _{env} = {\Omega _{free}} \cup {\Omega _{obstacle}} \in \mathbb{R}^{2}$, where $\Omega _{free}$ represents the free space for players' movement, $\Omega_{obstacle}$ containing static obstacles $\mathcal{O}_{s}$ and dynamic obstacles $\mathcal{O}_{d}$ represents the obstacle area. The defense group $\mathcal{M}_D = \{\mathcal{D}_L,\mathcal{D}_F\} $ involves a Leader Defender $\mathcal{D}_{L}=\left\{D_{0}\right\}$ and $N-1$ Follower Defender $\mathcal{D}_{F}=\left\{D_{1},D_{2},...,D_{N-1}\right\}$, their opponent is an attacker $A$ without strategy limitations. Suppose their kinematics satisfy the following equation of state:
\begin{equation}\label{kinematics function}
    \begin{array}{l}
        {{{\mathbf{\dot x}}}_A} = {{\mathbf{u}}_A} \,\quad {{\mathbf{x}}_A}(0) = {\mathbf{x}}_A^0\\
        {{{\mathbf{\dot x}}}_{Di}} = {{\mathbf{u}}_{Di}} \enspace {{\mathbf{x}}_{Di}}(0) = {\mathbf{x}}_{Di}^0 \ {i} \in I_d=\{0,1,...,N-1\}
    \end{array}
\end{equation}
where ${{{\mathbf{x}}}_A}$,${{{\mathbf{x}}}_{Di}}$ (${{\mathbf{u}}_A}$,${{\mathbf{u}}_{Di}}$)denotes the positions(control inputs) of Attacker and $i$th Defender respectively, and the top corner mark 0 denotes the value of the initial time, $I_d$ denotes the index set. Note that although we use the first-order integrator model in this paper to simplify the description of the problem, our method can be extended to systems of any order. Suppose there are velocity upper bounds $\left\lVert{{{\bf{\dot x}}}_A}\right\rVert\leqslant{\overline{V}_{A}}$ and $\left\lVert{{{\bf{\dot x}}}_{Di}}\right\rVert\leqslant{\overline{V}_{D}}$, and the bounds satisfy ${\alpha_{V}}={\frac{\overline{V}_{A}}{\overline{V}_{D}}}<1$, where $\alpha_{V}$ represents real speed ratio.\par

\subsection{Non-cooperative Indirect Guidance Framework}

\begin{figure*}[t]
    \centering
    \begin{minipage}[t]{0.45\textwidth}
        \centering
        \includegraphics[width=0.75\linewidth]{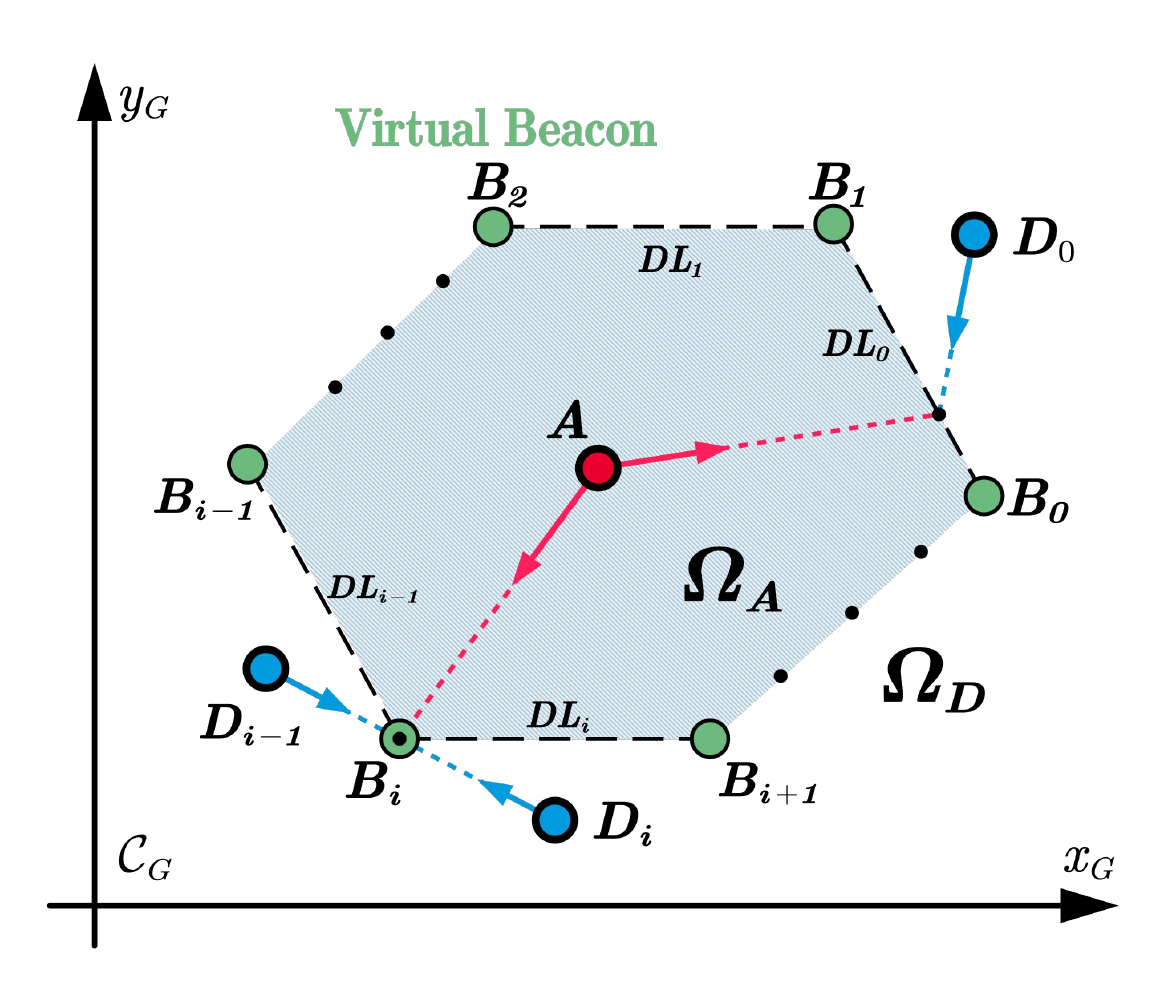}
        \subcaption{}
        \label{fig:sub1}
    \end{minipage}
    \hfill
    \begin{minipage}[t]{0.45\textwidth}
        \centering
        \includegraphics[width=0.75\linewidth]{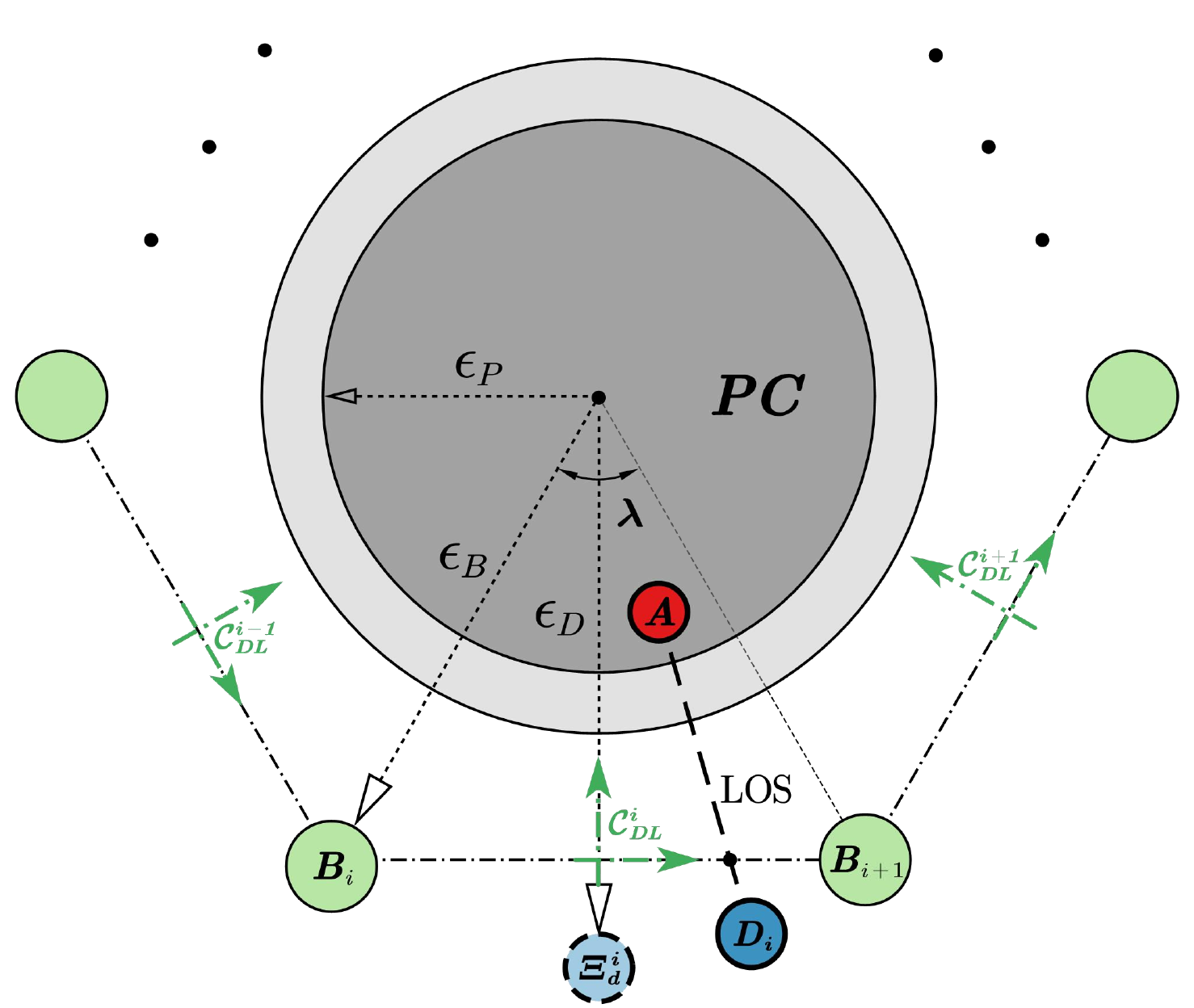}
        \subcaption{}
        \label{fig:sub2}
    \end{minipage}
    \caption{Diagram of the relative relationship among players and the coordinate system. }
    \label{fig:both}
\end{figure*}

To safely migrate attacker in a complex environment with dynamic obstacles, we propose a hybrid drive framework combining reach-avoid game and cooperative collision avoidance methods, and divide the non-cooperative indirect guidance problem into two subproblems called \textbf{capture problem} and \textbf{escort problem}, addressing them in capture stage and escort stage respectively. By introducing \textit{Pursuit Circle} $PC = \{\mathbf{r}_p \in \mathbb{R}^2 | \|\mathbf{r}_p-\mathbf{r}^k_{pc}\|\leqslant \epsilon_P \}$ and building an event-triggered update mechanism, an optimized initial distribution configuration $\mathcal{F}^{init}$ is designed for the defender group, which is of decisive significance to ensure the success of the escort task. We also define a regular N-sided enclosure polygon $\mathcal{E}_N = \{\mathcal{V_B},\mathcal{L_D}\}$ called \textit{Virtual Fence}, and it's bounded by virtual beacons $\mathcal{V_B} = \{B_i|i \in I_d\}$ as vertices and defense lines $\mathcal{L_D} = \{DL_i|i \in I_d\}$ as edges. Using the virtual fence as the boundary, $\Omega_{free}$ is further partitioned into an attacker-constrained space $\Omega_A$ within the polygon and a defender mobility space $\Omega_D$ outside the polygon.
\begin{definition}
    (Reachable Set):For a convex subset $\Omega_C \subset \Omega_{free}$ containing players, the set $\mathcal{R}^i_A=\{\mathbf{p} \in \mathbb{R}^2| \| \mathbf{x}_A-\mathbf{p}\Vert \leqslant \alpha_V\| \mathbf{x}_{Di}-\mathbf{p}\Vert \}$ describes the region in space where attacker $A$ can reach before defender $D_i$, The intersection $\mathcal{R}_A = \mathcal{R}^0_A \cap \mathcal{R}^1_A \cap ... \cap \mathcal{R}^{N-1}_A$ defines the combined region. When a point $\bf{p}$ lies within $\mathcal{R}_A$, there exists no feasible defense strategy allowing defenders to reach $\bf{p}$ faster than the attacker, i.e, the game situation is favorable to the attacker.
\end{definition}
\begin{definition}
    (Pursuit-Circle update mechanism): The pursuit circle describes a predefined region that allows the attacker to roam freely without invalidating the initial distribution configuration. At time $t_k$, an update is triggered when the attacker reaches the pursuit circle boundary, i.e., $\|\mathbf{x}_A(t_k)-\mathbf{r}^{k-1}_{pc}\| = \epsilon_P$, the circle radius $\epsilon_P$ remains constant, while its center position is updated to the attacker's current position, i.e., $\mathbf{r}^{k}_{pc} = \mathbf{x}_A(t_k)$.\par
\end{definition}

\begin{definition}
    (Initial distribution configuration): The initial distribution configuration $\mathcal{F}^{init}$ defines a set of desired positions $\{\mathbf{\Xi}^i_{D} \in \Omega_{free} |i \in I_d\}$ for defenders to suppress the attacker's escape intension at the beginning of escort stage. $\mathcal{F}^{init}$ remains synchronized with updates to the pursuit circle.
\end{definition}
\begin{remark}
    Because polygons are closed, in particular, When $i$ takes $N-1$, it is stipulated that $i+1$ is $0$.
\end{remark}
\begin{problem}
    (Capture Problem): Given a protected area $\mathcal{P}$ and pursuit circle radius $\epsilon_P$, assuming that one attacker starts from initial position $\mathbf{x}^0_A \in \Omega_{free}$ and $N$ defenders start from different initial positions $\{\mathbf{x}^0_{Di} \in \Omega_{free}|i \in I_d\}$, find a suitable control strategy $\{\mathbf{u}^{cap}_{Di} \in \mathbb{R}^2 | i \in I_d\}$ for every defender to ensure they can establish a closed formation that defined by initial distribution configuration $\mathcal{F}^{init}$ at $T_{f1}$ before $T_{ar}$, where $T_{ar}$ denotes the time that attacker reaches $\mathcal{P}$. We solve this problem in the capture stage.
\end{problem}
\begin{problem}
    (Escort Problem): Assuming that the defenders have formed $\mathcal{F}^{init}$, given a predefined target area $\mathcal{T}$, for any admissible attacker control input $\{\mathbf{u}_A \in \mathbb{R}^2 |\lvert\mathbf {u}_A\rvert \leqslant \overline{V}_A\}$, find collaborative motion strategies $\{\mathbf{u}^{ect}_{Di},\mathbf{v}_{Bi}\in \mathbb{R}^2| \lvert\mathbf{u}^{ect}_{Di}\rvert \leqslant \overline{V}_D, i \in I_d\}$ for every defender $D_i$ and virtual beacon $B_i$ such that: 1) for $\forall t \in [T_{f1},\infty)$, the attacker is confined inside the virtual fence $\mathcal{E}_N$ without touching defense lines $\mathcal{L_D}$, and all players move in $\Omega_{free}$; 2) the attacker can be driven into $\mathcal{T}$ at some time $T_{f2} \in (T_{f1},\infty)$.
\end{problem}

\section{Reachable Set Analysis}
We define the \textit{Game of Kind} as the problem of whether the attacker's reachable set can be constrained within the virtual fence at any moment during the escort stage. Therefore, this section will present the reachable set analysis and the game-theoretic situation assessment methods.
\begin{lemma}
    Within a convex domain $\Omega_C$, given a defender located at $\mathbf{x}_{D}=[x_d,y_d]^T$ and an attacker at $\mathbf{x}_A=[x_a,y_a]^T$, with a speed ratio of $\alpha_V$, the boundary of the attacker's reachable region $\mathcal{R}_A$ is an Apollonius Circle with center $\mathbf{r}_{ac}$, expressed as:
\begin{equation}
    \begin{aligned}\label{Apollonius Circle}
        {AC} = {\{\mathbf{r}_{a} \in \mathbb{R}^2 | \|\mathbf{r}_{a}-\mathbf{r}_{ac}\|=\frac{\alpha_V}{1-\alpha_V^2}\|\mathbf{x}_D-\mathbf{x}_A\|\}}
    \end{aligned}
\end{equation}
\end{lemma}

For the attacker $A$ and the defender $D_i$ that locate on the opposite sides of the defense line $DL_i$, let the connecting line between $A$ and $D_i$ be the Line of Sight (LOS), the vertical distances from $A$ and $D_i$ to $DL_i$ are $l^i_a$ and $l^i_d$, respectively, and the angle between the LOS and $DL_i$ is termed to be $\varphi^i_{los}$. Given the attacker's position $\mathbf{x}_A$, there exists a barrier curve $\mathcal{B}_D$ which divides the defender's free-move space $\Omega_D$ into $\mathcal{W}^A_D$(Attacker-win Region) and $\mathcal{W}^D_D$(Defender-win Region). When the defender lies on $\mathcal{B}_D$, the attacker's reachable set $\mathcal{R}_A$ and $DL_i$ share exactly one unique intersection point, i.e., they are tangent.

\begin{theorem}
     The analytical form of $\mathcal{B}_D$ can be expressed as follows:
\begin{equation}
    \begin{aligned}\label{Barrier Fcn}
        & {\mathcal{B}_D} = {\{\mathbf{x}_D \in \Omega_D | J(l^i_a,l^i_d,\varphi^i_{los})=0\}}\\
        & {J(l_a,l_d,\varphi_{los})=l_a(1-\frac{\alpha_V}{\sin{\varphi_{los}}})-l_d(\frac{\alpha_V}{\sin{\varphi_{los}}}-\alpha_V^2)}
    \end{aligned}
\end{equation}
where $J(l_a,l_d,\varphi_{los}):\mathbb{R}^3\to \mathbb{R}$ denotes the \textit{Judgment Function}, which reflects the advantageous states of the game.
\end{theorem}
\textit{Proof.} See Appendix A.\par
Note that $J>0 \Leftrightarrow D_i \in \mathcal{W}^D_D \Rightarrow \mathcal{R}_A \cap \mathcal{L}=\varnothing$ indicates the advantage of defender. Conversely, it turns the advantage of the attacker, i.e., $D_i \in \mathcal{W}^A_D \Rightarrow\mathcal{R}_A \cap \mathcal{L} \neq \varnothing$. Similarly, when the defender's position is given, we can also determine a barrier curve $\mathcal{B}_A$ for the attacker's free-move space $\Omega_A$ with the judgment function.

\begin{remark}
According to Theorem 1, for all $i$ in $I_d$, \textbf{two  necessary conditions} must be met simultaneously to guarantee defense victory
\begin{equation}
    \label{sufficient equations}
\begin{aligned}
    & J(l^i_d,l^i_a,\varphi^i_{los}) > 0 \Rightarrow \\
        & \begin{cases}
            \textbf{Condition 1}:
            |\varphi^i_{los}| \in [\pi/2-\cos^{-1}\alpha_V,\pi/2+\cos^{-1}\alpha_V]\\
            \textbf{Condition 2}:
            l^i_d < l^i_a[\frac{\sin\varphi^i_{los}-\alpha_V}{\alpha_V(1-\alpha_V\sin\varphi^i_{los})}]
        \end{cases}
\end{aligned}
\end{equation}
\end{remark}

In ground inertial cartesian coordinate frame $\mathcal{C}_{G}$, variable group $\{l^i_d,l^i_a,\varphi^i_{los}\}$ in \eqref{sufficient equations} can be obtained respectively through vector operations between defender $D_i$, attackers $A$ and adjacent beacons $DL_i=\{B_i,B_{i+1}\}$
\begin{equation}
    \begin{aligned}
        & l_d^i = \left| {\frac{{\left\langle {({\mathbf{x}}_{Di} - {\mathbf{\Xi}}^i_B),{\mathbf{T}} \cdot ({\mathbf{\Xi}}_B^{i + 1} - {\mathbf{\Xi}}_B^i)} \right\rangle }}{{\left\| {{\mathbf{\Xi}}_B^{i + 1} - {\mathbf{\Xi}}_B^i} \right\|}}} \right|\\
        & l_a^i = \left| {\frac{{\left\langle {({{\mathbf{x}}_A} - {\mathbf{\Xi}}_B^i),{\mathbf{T}} \cdot ({\mathbf{\Xi}}_B^{i + 1} - {\mathbf{\Xi}}_B^i)} \right\rangle }}{{\left\| {{\mathbf{\Xi}}_B^{i + 1} - {\mathbf{\Xi}}_B^i} \right\|}}} \right|\\
        & \varphi^i_{los}= \pi-\cos^{-1} (\left\langle {\frac{{{{\mathbf{x}}_A} - {\mathbf{x}}_{Di}}}{{\left\| {{{\mathbf{x}}_A} - {\mathbf{x}}_{Di}} \right\|}},\frac{{{\mathbf{\Xi}}_B^{i + 1} - {\mathbf{\Xi}}_B^i}}{{\left\| {{\mathbf{\Xi}}_B^{i + 1} - {\mathbf{\Xi}}_B^i} \right\|}}} \right\rangle )\\
        & {\mathbf{T}} = \left[ {\begin{array}{*{20}{c}}
  0&{ - 1} \\ 
  1&0 
\end{array}} \right]
    \end{aligned}
\end{equation}
where $\langle \mathbf{a},\mathbf{b} \rangle $ denotes the vector inner product between $\textbf{a}$ and $\textbf{b}$, matrix $\mathbf{T}$ rotates the direction of $\overrightarrow{B_{i}B_{i+1}}$ clockwise by $\pi/2$ to obtain the normal vector of $DL_i$.

\section{Control Strategy}
\subsection{Capture Stage}
To solve \textit{Problem 1} in capture stage, we first define the basic expression of $\mathcal{F}^{init}$ and $\mathcal{E}_N$. As shown in Fig.\ref{fig:both}, since virtual fence $\mathcal{E}_N$ is an N-side regular polygon (Let $\lambda = 2\pi/N$), and defender $D_i$ aims to guard $DL_i$, according to symmetry, it is reasonable to evenly place $\mathbf{\Xi}^i_{D}$ on the perpendicular bisector of $DL_i$, so $\mathcal{F}^{init}$ is also an N-side regular polygon. Let $k$ be the current $PC$ update counter, and $i$ be the defender index. In $\mathcal{C}_G$, we can easily get $\mathbf{\Xi}^i_{D} = \mathbf{r}^k_{pc}+\epsilon_D \cdot [\cos\theta^i_d,\sin\theta^i_d]^T$, where $\epsilon_D=\|\mathbf{\Xi}^i_{D}-\mathbf{r}^k_{pc}\|$, and $\theta^i_d = (i+\frac{1}{2}) \cdot \lambda$ denotes the distribution angle. Similarly, the position of virtual beacon $B_i$ can be expressed as $\mathbf{\Xi}^i_{B} = \mathbf{r}^k_{pc}+\epsilon_B \cdot [\cos\theta^i_b,\sin\theta^i_b]^T$, where $\theta^i_b = i \cdot \lambda$.\par
We notice that \textbf{Condition 1} is solely related to $\epsilon_D$, and once $\epsilon_D$ is determined, \textbf{Condition 2} depends only on $\epsilon_B$. Therefore, our algorithm decouples the design task of $\mathcal{F}^{init}$ and computes $\epsilon_D$ and $\epsilon_B$ sequentially.

\begin{theorem}
Given the $PC$ radius $\epsilon_P$ and fence sides number $N$, exist optimal parameter combination $\{\epsilon_D,\epsilon_B\}$, which ensures that $\mathcal{F}^{init}$ simultaneously satisfies \eqref{sufficient equations}
\begin{subequations}\label{formation param calculation}
    \begin{align}
        & \epsilon_D = 
        \begin{cases}
            K_p\epsilon_P[\frac{\sin(\lambda/2)\hat{\alpha_V}}{\sqrt{1-\hat{\alpha_V}^2}}+\cos(\frac{\lambda}{2})] &\Psi > \frac{\lambda}{2}\\
            K_p\epsilon_P\frac{1}{\sqrt{1-\hat{\alpha_V}^2}} &\Psi \leqslant \frac{\lambda}{2}
        \end{cases}\label{formation param calculation.1}\\
        & \epsilon_B = \begin{cases}
            \min[\, f(0), f(\eta^*), f(\lambda/2)] &\eta* \in[0,\lambda/2]\\
            \min[\, f(0), f(\lambda/2)] &\eta* \notin [0,\lambda/2]
        \end{cases}\label{formation param calculation.2}\\
        & \Psi = \sin^{-1}(\hat{\alpha_V}) = \frac{\pi}{2} - \bar{\Psi}\label{formation param calculation.3}\\
        & \begin{aligned}
            f(\eta) = &\frac{1}{1-\hat{\alpha_V}^2}(\epsilon_D-\epsilon_P\cos{\eta}\\
            & -\hat{\alpha_V}\sqrt{\epsilon_P^2+\epsilon_D^2-2\epsilon_P\epsilon_D\cos{\eta}})\\
        \end{aligned}\\
        & \eta* = \cos^{-1}\frac{[\epsilon_P^2+(1-\hat{\alpha_V}^2)\epsilon_D^2]}{2\epsilon_P\epsilon_D} = \cos^{-1}\chi
    \end{align}
\end{subequations}
\end{theorem}
\textit{Proof.} See Appendix B.\par
\eqref{formation param calculation} provide the calculation method of $\{\epsilon_D,\epsilon_B\}$, where $\bar{\Psi}=\cos^{-1}{\hat{\alpha_V}}$ denotes the upper bound of lateral angle absolute error $|e^i_{\varphi}|=|\pi/2-\varphi^i_{los}|$, $K_p > 1$ denotes the zoom factor, which is used to avoid geometric singularity problem, $\hat{\alpha_V} \in (\alpha_V,1)$ is a user-defined parameter of providing margins for migration speed which will be discussed later.\par
$\mathcal{F}^{init}$ turns the complex \textit{Problem 1} into simpler \textit{Task-assignment Problem} and \textit{Position-tracking Problem} between $\{\mathbf{\Xi}^i_D|i \in I_d\}$ and $\{\mathbf{x}_{Di}|i \in I_d\}$. \cite{chipadeMultiagentPlanningControl2021} modeled the \textit{Task-assignment Problem} as a generalized assignment problem (GAP), and developed a mixed integer quadratically constrained program(MIQCP) to find the near-optimal assignment. Let $\mathbf{P}^j_i$ be the shortest path from $\mathbf{x}_{Di}(t_k)$ to $\mathbf{\Xi}^{j}_{D}$, and $\mathbf{P}^{j'}_{i'}$ be the shortest path from $\mathbf{x}_{Di'}(t_k)$ to $\mathbf{\Xi}^{j'}_{D}$, both $\mathbf{P}^j_i$ and $\mathbf{P}^{j'}_{i'}$ are $\mathcal{C}^1$ paths that bypasses convex obstacles in a tangential manner. Let $T^{\Xi,j}_{D,i}$ denotes the time that $\mathbf{P}^j_i$ will take without considering possible collusion between trajectories, $S^{i',j'}_{i,j}$ denotes the length of the intersection section between $\mathbf{P}^j_i$ and $\mathbf{P}^{j'}_{i'}$, MIQCP can be formulated as:
\begin{equation}
    \begin{aligned}
        \mathrm{minimize} \quad &{\Sigma T^{\Xi,j}_{d,i} a_{ij} + \Sigma S^{i',j'}_{i,j} a_{ij}a_{i'j'}}\\
        \mathrm{subject \, to} \quad &\Sigma_{j \in I_d} a_{ij}=1 \;\; \forall i \in I_d \\
        & \Sigma_{j \in I_d} a_{ij}=1 \;\; \forall j \in I_d\\
        & a_{i',j'},a_{i,j} \in \{0,1\} \;\; \forall i',j',i,j, \in I_d
    \end{aligned}
\end{equation}
when $a_{ij} = 1$, $D_i$ is assigned to $\mathbf{\Xi}^j_{d}$, otherwise $a_{ij} = 0$. Note: For all $i,j \in I_d$, we can rename the subscript of $D_i$ to make it the same as the subscripts of the corresponding target point $\mathbf{\Xi}^j_d$, which will not cause any problems.\par
Since defenders and their goals have been one to one paired in MICQP, and $\mathcal{F}^{init}$ is stationary within one update cycle, \textit{Position-tracking Problem} can be solved with single-unit obstacle avoidance algorithm, such as Artificial Potential Field (APF) and Vector Field Histogram (VFH). In this article we use APF, with  potential function defined as following
\begin{equation}\label{single potential function}
    \begin{aligned}
        & {{U^{cap}_{a,i}}(\mathbf{x}_{Di},\mathbf{\Xi^i_D})} = {\frac{1}{2}K^{cap}_{a}{d^{\Xi,i}_{D,i}}^2}\\
        & {{U^{cap}_{r,i,k}}(\mathbf{x}_{Di},O_k)} = \begin{cases}
            \frac{1}{2}{K^{cap}_{r}}(\frac{1}{{d_{D,i}^{O,k}}} - \frac{1}{\Gamma^{cap}})^2 \quad &d_{D,i}^{O,k} < \Gamma^{cap}\\
            0 &\textbf{Otherwise}
        \end{cases}
    \end{aligned}
\end{equation}
where $d^{\Xi,i}_{d,i} = \left\| {{\mathbf{x}}_{Di} - {{\mathbf{\Xi}}^i_D}} \right\|$, $d_{D,i}^{O,k} = \left\| {{\mathbf{x}}_{Di} - {\mathbf{p}}_{min}^k} \right\| = \mathop {\min }\limits_{{{\mathbf{p}}} \in \partial {\mathcal{O}_k}} \left\| {{\mathbf{x}}_{Di} - {{\mathbf{p}}}} \right\|$, $K^{cap}_{a,i}$ and $K^{cap}_{r,i}$ are control gains, $\partial \mathcal{O}_k$ denotes the boundary of $O_k$, $d^{cap}_{\tau}$ is obstacle avoidance threshold. The total potential value at point $\mathbf{x}_{Di}$ is $U^{cap}_i = U^{cap}_{a,i} + \Sigma^{m}_{k=1} U^{cap}_{r,i,k}$, we get the virtual force $\mathbf{F}^{cap}_i = -\nabla U^{cap}_i$
\begin{equation}
    \begin{aligned}
        & \mathbf{F}^{cap}_{a,i} = {K^{cap}_{a}}({{\mathbf{\Xi}}^i_D} - {\mathbf{x}}_{Di}) \\
        & \mathbf{F}^{cap}_{r,i,k} = \begin{cases}
            {{K^{cap}_{r}}[(\frac{1}{{d_{D,i}^{O,k}}} - \frac{1}{\Gamma^{cap} })\frac{{({\mathbf{x}}_{Di} - {\mathbf{p}}_{min}^k)}}{{d_{D,i}^{O,k}}^3}}] \quad &d_{D,i}^{O,k} < \Gamma^{cap}\\
            0 & \textbf{Otherwise}
        \end{cases}
    \end{aligned}
\end{equation}

Furthermore, in order to avoid collisions among defenders, the internal exclusion term is defined as
\begin{equation}
    \begin{aligned}
        \mathbf{F}^{int}_{i,j} = \begin{cases}
            K^{int}(\frac{1}{\varrho_{ij}}-\frac{1}{\Gamma^{int}})(\textbf{x}_{Di}-\textbf{x}_{Dj}) \quad & \varrho_{ij} < \Gamma^{int} \\
            0 & \textbf{Otherwise}
        \end{cases}
    \end{aligned}
\end{equation}
where $K^{int}$ is a positive gain coefficient, $\varrho_{ij} = \|\textbf{x}_{Di}-\textbf{x}_{Dj}\|$ denotes the relative distance between defender $i$ and defender $j$. For this, the total force $\textbf{F}^{cap}_{i}$ and control input $\mathbf{u}^{cap}_{Di}$ of $i$th defender are set as
\begin{subequations}
\begin{align}
        & \textbf{F}^{cap}_{i} = \mathbf{F}^{cap}_{a,i} + \sum^{m}_{k=1}\mathbf{F}^{cap}_{r,i,k}+\sum^{N-1}_{j = 0, j \neq i}\mathbf{F}^{int}_{i,j} \label{subeq:cap total force}\\
        & \mathbf{u}^{cap}_{Di} = \Omega_{\bar{V}_D}(\textbf{F}^{cap}_{i}) \label{subeq:cap input}
\end{align}
\end{subequations}
Where $\Omega_{\bar{u}}(\mathbf{a})=\mathrm{min}(\|\mathbf{a}\|,\bar{u})\frac{\mathbf{a}}{\|\mathbf{a}\|}:\mathbb{R}^2 \rightarrow \mathbb{R}^2$ is a saturation function, which makes sure that $\|\mathbf{u}^{cap}_{Di}\| \leqslant \bar{V}_D$.
\subsection{Escort Stage}
\subsubsection{Game Layer}
By regarding $\mathcal{L}_D$ as the target line, we turn \textit{Problem 2} into the TAD (Target-Attack-Defense) problem alone each edge of $\mathcal{E}_N$. As shown in Fig \ref{fig:sub1}, the conflict between $A$ and  $D_i$ is termed the $1 \ vs. \ 1$ static reach-avoid sub-game in the defense line cartesian coordinates $\mathcal{C}^i_{DL}$. Still consider the sufficient conditions, we formulate the horizontal and vertical control targets separately. The horizontal control target restricts the LOS error angle $e^i_{\varphi}$ within the range required by condition 1, and the vertical control target restricts the vertical distance $l^i_d$ within the range required by condition 2. Let the desired value of $\varphi_{los}$ be $\varphi_D = \pi/2$, the desired value of $l^i_d$ be $\ell_D^i$, we establish the error form of \eqref{sufficient equations}:
\begin{subequations}\label{Rewrite}
    \begin{flalign}
        & \forall t \in [T_{f1},\infty),\, \forall i \in I_d, \ \text{s.t.} 
        \nonumber\\
        & \left| e_h^i \right| = \left| \left\langle \mathbf{x}_A - \mathbf{x}_{Di},\ 
           \frac{\mathbf{\Xi}_B^{i+1} - \mathbf{\Xi}_B^i}{\left\| \mathbf{\Xi}_B^{i+1} - \mathbf{\Xi}_B^i \right\|} 
           \right\rangle \right| < (l_d^i + l_a^i)\tan(\bar{\Psi})\label{rewrite_1}\\
        & \Rightarrow \textbf{Condition 1}
        \nonumber\\
        & -(1-K_{\Delta}) \frac{\cos(e_\varphi^i) - \hat{\alpha}_V}{1 - \cos(e_\varphi^i)\hat{\alpha}_V} \frac{l^i_a}{\hat{\alpha}_V} < e_v^i = l_d^i - \ell_D^i \label{rewrite_2}\\
        & < K_{\Delta}\frac{\cos(e_\varphi^i) - \hat{\alpha}_V}{1 - \cos(e_\varphi^i)\hat{\alpha}_V} \frac{l^i_a}{\hat{\alpha}_V}\nonumber\\
        & \Rightarrow \textbf{Condition 2}
        \nonumber
    \end{flalign}
\end{subequations}
where $e_h^i$ and $e_v^i$ are control error quantities, and the right side of the inequalities give the tolerance range, let $\Delta = K_{\Delta}l^i_a \in (0,l^i_a)$ be an adaptive factor, and we design $\ell_D^i$ as
\begin{equation}\label{ell_D}
    \ell _D^i \triangleq \frac{{l_a^i - \Delta }}{{{{\hat{\alpha_V}}}}}[\frac{{\cos (e_\varphi ^i) - {{\hat{\alpha_V}}}}}{{1 - \cos (e_\varphi ^i){{\hat{\alpha_V}}}}}]
\end{equation}

Since $D_i$ and $A$ are located on either side of $DL_i$ respectively, and $l_d^i,l_a^i>0$, it can be easily obtained from the triangular geometric relationship that \eqref{rewrite_1} fully satisfies \textbf{Condition 1}. Substitute \eqref{ell_D} back into \eqref{rewrite_2}, we get $0<l_d^i<\frac{{l_a^i}}{{{{\hat{\alpha_V}}}}}[\frac{{\cos(e_\varphi ^i) - {{\hat{\alpha_V}}}}}{{1 - \cos (e_\varphi ^i){{\hat{\alpha_V}}}}}]$, so \textbf{Condition 2} can also be fully deduced from \eqref{rewrite_2} and \eqref{ell_D}. At this point, we have transformed the TAD problem into a constrained control problem.

\begin{lemma}\cite{bechlioulisRobustAdaptiveControl2008a}
    In nonlinear system $\mathbf{\dot x} = \mathbf{F(x)}+\mathbf{G(x)u}$ with the initial tracking error of $x_i$ satisfying $0<e_i(0)<\rho_{i,0}<\bar{e_i}$, and a Prescribe Performance Function is defined as $\rho_i(t) = (\rho_{i,0}-\rho_{i,\infty})\exp(-\kappa t)+\rho_{\infty}$($0<\rho_{i,\infty}<\rho_{i,0}$), the constrained control problem $-\delta_{i}\rho_i(t)<e(t)<\rho_i(t)(0 \leqslant \delta_{i} \leqslant 1)$ can be transformed into the stable control problem of $\varepsilon$ through \textit{Error Transformation Function} $S(\varepsilon_i)=\frac{e_i}{\rho_i}:\mathbb{R} \rightarrow \mathbb{R}$ with the following properties:
\begin{align}
    \text{1)}: & \ S(\varepsilon) \text{ is smooth and strictly increasing;} \\
    \text{2)}: & \ \begin{aligned}[t]\label{property 2}
                  \lim_{\varepsilon_i \to -\infty} S(\varepsilon_i) &= -\delta_i \\
                  \lim_{\varepsilon_i \to \infty} S(\varepsilon_i) &= 1
                \end{aligned}
\end{align}
similarly, when $-\rho_{i,0} < e_i(0) < 0$, prescribe performance function is $-\rho_i(t)<e(t)<\delta_{i}\rho_i(t)$, and property \eqref{property 2} is turned to $\lim_{\varepsilon_i \to -\infty} S(\varepsilon_i) = -1, \lim_{\varepsilon_i \to \infty} S(\varepsilon_i) = \delta_{i}$.
\end{lemma}

At the initial moment $T_{f1}$ of the escort stage, taking the case that $e_v^i(T_{f1}) > 0$ as an example, then we define $\delta^f_i = \min(\frac{1-K_{\Delta}}{K_{\Delta}},1)$, $\delta^g_i = 1$, \eqref{rewrite_2} can be rewritten as
\begin{subequations}
    \begin{flalign}
       & -\delta^g_i < \widetilde{e_h^i}  = \frac{e_h^i}{\widetilde{g}} < 1
       \label{normalization err 1}\\
       & -\delta^f_i < \widetilde{e_v^i} = \frac{e_v^i}{\widetilde{f}} < 1
       \label{normalization err 2}
    \end{flalign}
\end{subequations}
where $\widetilde{g} \triangleq (l_d^i + l_a^i)\tan(\bar{\Psi})$ and $\widetilde{f} \triangleq \frac{\cos(e_\varphi^i) - \hat{\alpha}_V}{1 - \cos(e_\varphi^i)\hat{\alpha}_V} \frac{K_{\Delta}l^i_a}{\hat{\alpha}_V}$ are normalization functions, $ \widetilde{e_h^i}, \widetilde{e_v^i} \in (-1,1)$ are relative errors. \par
Define the Prescribe Performance Function as
\begin{equation}
    \begin{array}{l}
        \rho(t) = (1-K_{\infty})\exp(-\kappa t)+K_{\infty}\\
        K_{\infty} \in (0,1)
    \end{array}
\end{equation}
where $t$ is the time elapsed starting from $T_{f1}$, $\kappa$ is a positive coefficient. Define the Error Transformation Function as
\begin{equation}\label{error transform function}
        S(\varepsilon) = \frac{\exp(\varepsilon)-\delta\exp(-\varepsilon)}{\exp(\varepsilon)+\delta\exp(-\varepsilon)} = \frac{\widetilde{e}}{\rho}
\end{equation}
then the expected transient normalization error is constrained within $-\delta^g_i \rho(t)< \widetilde{e_h^i} < \rho(t)$ and $-\delta^f_i \rho(t)< \widetilde{e_v^i} < \rho(t)$.
\begin{remark}
    While $e_h^i(T_{f1}) < 0$, then take $\delta^f_i = \min(\frac{K_{\Delta}}{1-K_{\Delta}},1)$ and $\widetilde{f} = \frac{\cos(e_\varphi^i) - \hat{\alpha}_V}{1 - \cos(e_\varphi^i)\hat{\alpha}_V} \frac{(1-K_{\Delta})l^i_a}{\hat{\alpha}_V}$, as illustrated in Lemma 2, \eqref{normalization err 2} and \eqref{error transform function} are changed to
\begin{subequations}
    \begin{flalign}
        & -1 < \widetilde{e_v^i} = \frac{e_v^i}{\widetilde{f}} < \delta^f_i\\
        & S(\varepsilon) = \frac{\delta\exp(\varepsilon)-\exp(-\varepsilon)}{\delta\exp(\varepsilon)+\exp(-\varepsilon)}\label{error transform function 2}
    \end{flalign}
\end{subequations}
\end{remark}

Taking $\mathbf{g}^i_{DL} = [g^i_h,g^i_v]^T$ as the control components of $D_i$ in $\mathcal{C}^i_{DL}$, we also notice that when $K_{\Delta}=\frac{1}{2}$, both $\frac{1-K_{\Delta}}{K_{\Delta}}$ and $\frac{K_{\Delta}}{1-K_{\Delta}}$ are equal to $1$, so $S(\varepsilon)$ and $\widetilde{f}$ can be unified as $S(\varepsilon) = \frac{\exp(\varepsilon)-\exp(-\varepsilon)}{\exp(\varepsilon)+\exp(-\varepsilon)}=\tanh(\varepsilon)$ and $\widetilde{f}=\frac{l^i_a}{2\hat{\alpha}_V}\frac{\cos(e_\varphi^i) - \hat{\alpha}_V}{1 - \cos(e_\varphi^i)\hat{\alpha}_V} $, then $\delta^f_i = 1$ and $\dot \varepsilon$ has the simplest form under this condition. With $\dot\varepsilon = \frac{\partial{S}^{-1}(\widetilde{e}/\rho)}{\partial (\widetilde{e}/\rho)}\frac{d(\widetilde{e}/\rho)}{dt}$, the transformed error state equations of horizontal and vertical channels are obtained as
\begin{subequations}\label{transformed system}
    \begin{flalign}
        & \dot{\varepsilon}^i_h = \frac{\dot{e}^i_h(\widetilde{g}\rho) - e^i_h(\dot{\widetilde{g}} \rho + \widetilde{g} \dot{\rho})}{(\widetilde{g} \rho)^2-{e^i_h}^2}\label{horizontal function}\\
        & \dot{\varepsilon}^i_v = \frac{\dot{e}^i_v(\widetilde{f}\rho) - e^i_v(\dot{\widetilde{f}} \rho + \widetilde{f} \dot{\rho})}{(\widetilde{f} \rho)^2-{e^i_v}^2}\label{vertical function}
    \end{flalign}
\end{subequations}\par
Our goal is to ensure the transformed system \eqref{transformed system} completely stable, which means $\lim_{t \rightarrow \infty}\varepsilon = 0$.

\begin{lemma}
    Let Lyapunov Function $V(x)$ be a scalar function with continuous first partials satisfying
\begin{equation}
    \begin{aligned}
        & V(x)>0 \quad forall \quad x \neq 0\\
        & \dot{V}(x)<0 \quad forall \quad x\\
        & V(x) \rightarrow \infty \quad as \quad x \rightarrow \infty
    \end{aligned}
\end{equation}
then the system is completely stable.
\end{lemma}

Take the vertical control channel as example, we define the Lyapunov Function as $V(\varepsilon^i_{v}) = \frac{1}{2} {\varepsilon^i_{v}}^2$, and the first-order time derivative of $V(\varepsilon^i_{v})$ is obtained as $\dot V(\varepsilon^i_{v}) = -\varepsilon^i_{v} \dot{\varepsilon^i_{v}}$. When $\dot \varepsilon^i_{v}$ satisfies
\begin{equation}\label{Lyapunov prop 2}
    \dot \varepsilon^i_{v} = -K_v\varepsilon^i_{v} \ ,K_v>0
\end{equation}\par
It is clear that $\dot V(\varepsilon^i_{v})=-K_v{\varepsilon^i_v}^2<0$, so according to Lemma 3, the vertical control channel is completely stable. By substituting $\dot e^i_v = g^i_v -(- \dot \ell^i_D)$ and \eqref{Lyapunov prop 2} into \eqref{vertical function}, we have the vertical control component $g^i_v$
\begin{equation}
    g^i_v = \frac{-K_v \varepsilon^i_v[(\widetilde{f} \rho)^2-{e^i_v}^2]+e^i_v(\dot{\widetilde{f}} \rho + \widetilde{f} \dot{\rho})}{\widetilde{f} \rho} - \dot{\ell}^i_D
\end{equation}\par
Similarly, the horizontal control component can be derived as 
\begin{equation}
    g^i_h = \frac{-K_h \varepsilon^i_h[(\widetilde{g} \rho)^2-{e^i_h}^2]+e^i_h(\dot{\widetilde{g}} \rho + \widetilde{g} \dot{\rho})}{\widetilde{g} \rho} + \dot{x}^{DL}_{a,i}
\end{equation}
where $\dot{x}^{DL}_{a,i}$ represents the velocity component of $A$ alone the $DL_i$ direction. Note that under the actual discrete computing conditions, both continuous derivative terms $\dot{x}^{DL}_{a,i}$ and $\dot{\ell}^i_D$ can be replaced by the first-order difference of the observed values with respect to time. \par
Further, we map $\mathbf{g}^i_{DL}$ to the ground coordinate $\mathcal{C}_G$ through the rotation matrix $\mathbf{T}_{DL,i}^G$, and consider the upper bound constraint of the defender's speed, then the control input $\mathbf{u}^{ect}_{Di}$ under $\mathcal{C}_G$ can be expressed as
\begin{subequations}\label{game strategy}
    \begin{flalign}
        & \mathbf{u}^{ect}_{Di} = \Omega_{\bar{V}_D}({\mathbf{T}}_{DL,i}^G\mathbf{g}^i_{DL})\\
        & {\mathbf{T}}_{DL,i}^G = \left[ {\begin{array}{*{20}{c}}
  {\cos {\theta _i}}&{\sin {\theta _i}} \\ 
  { - \sin {\theta _i}}&{\cos {\theta _i}} 
\end{array}} \right]
    \end{flalign}
\end{subequations}
where $\theta_i$ denotes the rotation angle between $\mathcal{C}^i_{DL}$ and $\mathcal{C}_{G}$.
\subsubsection{Plan Layer}
The Plan Layer conducts local motion planning for beacon group $\mathcal{V}_B$ through \textit{Joint APF Approach}, thereby guiding the defender group $\mathcal{M}_D$ to achieve collaborative obstacle avoidance. To ensure the consistency and synchronization, we apply the Leader-Follower communication topology. The planning task is only deployed on Leader Defender $\mathcal{D}_L$, and as a defense node, $\mathcal{D}_L$ also needs to collect environmental information. The Follower Defenders $\mathcal{D}_F$ are responsible for uploading environmental detection information to $\mathcal{D}_L$ and receiving the planning action given by $\mathcal{D}_L$. Besides, in this article, we make the virtual fence have only translational movement and no rotational movement to simplify the analysis, thus all beacons share the same strategy, i.e., for all $i$ in $I_d$, $\mathbf{v}_{Bi} = \mathbf{v}_{Fc} = [v^x_{Fc},v^y_{Fc}]^T$, where $\mathbf{v}_{Fc}$ denotes the velocity of the center of virtual fence.\par
Let $\mathbf{x}_{Tc}$ denote the position of a certain point inside $\mathcal{T}$, similar to \eqref{single potential function}, we define the single-unit virtual attractive potential function $U^{ect}_{Ta,i}$ and repulsive potential function $U^{ect}_{Pr,i}$ and $U^{ect}_{r,i}$ for $D_i$ as
\begin{equation}
    \begin{aligned}
        & {{U^{ect}_{Ta,i}}(\mathbf{x}_{Di},\mathbf{x}_{Tc})} = {\frac{1}{2}K^{ect}_{Ta}{d^{T}_{D,i}}^2}\\
        &  {U^{ect}_{Pr,i}}(\mathbf{x}_{Di},\mathbf{x}_{Pc}) = -\frac{1}{2}K^{ect}_{Sr}\frac{1}{{d^{P}_{D,i}}^2} \\
        & {{U^{ect}_{r,i,k}}(\mathbf{x}_{Di},O_k)} = \begin{cases}
            \frac{1}{2}{K^{ect}_{r}}{(\frac{1}{{d_{D,i}^{O,k}}} - \frac{1}{{{\Gamma^{ect}}}})^2} \quad &d_{D,i}^{O,k} < \Gamma^{ect}\\
            0 &\textbf{Otherwise}
        \end{cases}
    \end{aligned}
\end{equation}
where $K^{ect}_{Ta}$, $K^{ect}_{Pr}$and $K^{ect}_{r}$ are positive gain coefficients, $\Gamma^{ect}$ is obstacle avoidance threshold, $d^{T}_{D,i} = \|\mathbf{x}_{Tc} - \mathbf{x}_{Di}\|$ and $d^{P}_{D,i} = \|\mathbf{x}_{Pc} - \mathbf{x}_{Di}\|$ respectively represent the relative distance from defender $D_i$ to the safe area and protected area. The virtual force $\mathbf{F}^{ect}_i$ felt by $D_i$ is expressed as
\begin{equation}
    \mathbf{F}^{ect}_i = -\nabla({U^{ect}_{a,i}}+U^{ect}_{Pr,i}+\sum^{m}_{k=1}{U^{ect}_{r,i,k}})
\end{equation}\par
Regard the defender group $\mathcal{M}_D$ as a whole, the resultant force $\mathbf{F}^{ect}_{\Sigma}$ acting on it is the superposition of the virtual forces experienced by each individual defender, which integrates the environmental information detected by all defenders
\begin{equation}
    \mathbf{F}^{ect}_{\Sigma} = \sum^{N-1}_{i=0}\mathbf{F}^{ect}_i
\end{equation}\par
Although the beacons' movement will bring additional disturbances to the game layer which is based on static analysis, by setting a reasonable upper bound of the beacon movement speed, we can still ensure the effectiveness of the distribution configuration $\mathcal{F}^{init}$ and the convergence of the control law $\mathbf{u}^{ect}_{Di}$.

\begin{theorem}
    For defender and attacker with upper speed bound $\overline{V}_D$ and $\overline{V}_A$, given the design speed ratio $\hat{\alpha_V} \in (\frac{\overline{V}_A}{\overline{V}_D},1)$, when $\|\mathbf{v}_{Fc}\|$ is restricted to below upper bound $\overline{V}_B$
\begin{equation}
    \overline{V}_B \leqslant \min(\frac{\hat{\alpha_V}\overline{V}_D-\overline{V}_A}{1+\hat{\alpha_V}},\overline{V}_A)
\end{equation}
the defense configuration and strategy defined by \eqref{formation param calculation} and \eqref{game strategy} are still valid.
\end{theorem}
\textit{Proof.} See Appendix C.\par

As the direction is guided by the virtual resultant force and the speed magnitude is limited by Theorem 3, the movement strategy for beacons is therefore expressed as
\begin{equation}
    \mathbf{v}_{Fc} = \Omega_{\overline{V}_B}(\mathbf{F}^{ect}_{\Sigma})
\end{equation}
\begin{remark}
    (Extension to Higher-Order Systems) The proposed framework is not restricted to the single-integrator model; equation \eqref{kinematics function} can be generalized to accommodate more complex robotic systems with higher-order nonlinear dynamics, such as quadrotors. As demonstrated in \cite{bechlioulisRobustAdaptiveControl2008a}, the constraint formulation in \eqref{Rewrite} remains valid, and the transformed error system analogous to \eqref{transformed system} can always be constructed. The corresponding stabilization controller can then be systematically designed using the backstepping approach.
\end{remark}

\section{Numerical Simulation}
In this section, numerical simulation is performed to verify the effectiveness of the proposed strategies, providing results for both the capture stage and the escort stage. In capture stage, the attacker try to enter the protected area, while the defender group sets out from an initial position around the protected area to search for the attacker and try to surround it, then enters the escort stage after forming $\mathcal{F}^{init}$. In escort stage, we assume that the attacker's policy consists of two parts: 1) When any defender enters the escape range, it evades the defenders through the APF method, 2) it attempts to break through the encirclement by adopting a random motion strategy when the evasion is not triggered, so that the defenders cannot know the attacker's movement in advance.
\subsection{Parameter and Environment Set}
The proposed method was evaluated using a computer with a Core AMD R5-4500H processor and 16 GB of memory. In our simulation, the maximum speeds of the attacker and the defender are set to $1.2(m/s)$ and $3.0(m/s)$, respectively. Not losing generality, the attacker starts from the map origin $[0,0]^T$, and its escape range $R_e$ is set to $0.8m$. $\Gamma^{cap}$ and $\Gamma^{ect}$ are both set to $8m$. The control period is $0.05s$. The principal parameters are detailed in Table \ref{tab:parameters}.
\begin{table}[!ht]
\centering
\caption{Parameters and values}
\label{tab:parameters}
\begin{tabular}{lcp{4cm}}
\toprule
Symbol & Value & Description      \\ 
\midrule
$N$ & 3 & Number of defenders\\
$\epsilon_p$ & $0.5 (m)$ & Radius of Pursuit Circle \\
$\hat{\alpha}_V$ & $0.65$ & Design speed ratio \\
$K_p$ & $2$ & Expansion factor \\
$\kappa$ & 1 & Convergence coefficient \\
$K_{\infty}$ & 0.8 & Terminal error constraint \\
$\mathbf{x}_{Pc}$ & $[5,20]^T (m)$ & Position of protected area \\
$\mathbf{x}_{Tc}$ & $[20,20]^T (m)$ & Position of safe area \\
\bottomrule
\end{tabular}
\end{table}
\subsection{Result Analysis}
\begin{figure}[!htbp]
    \centering
    \begin{minipage}{0.48\textwidth}
        \centering
        \includegraphics[width=0.7\linewidth]{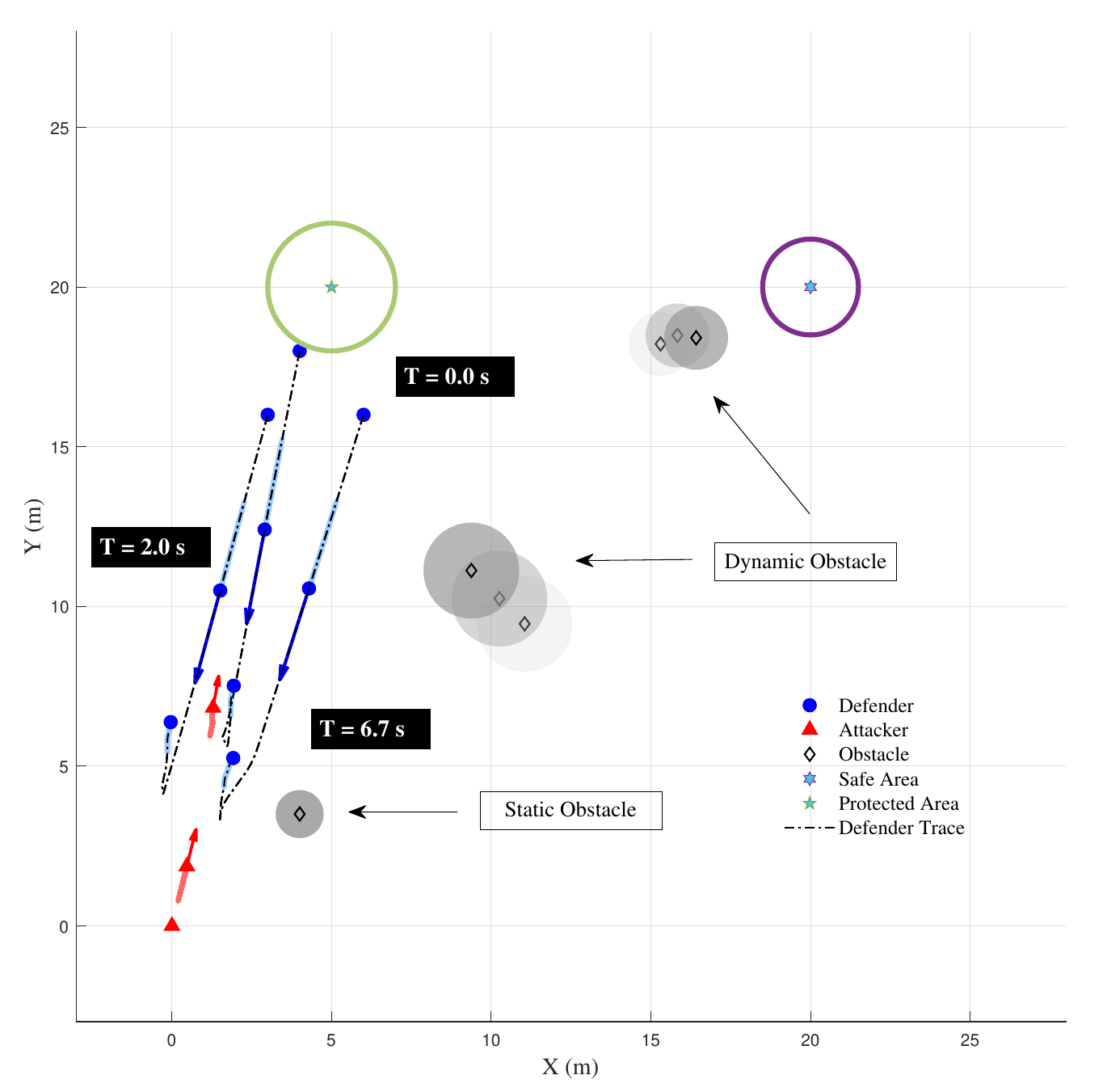}
        \subcaption{}
        \label{fig:stage1}
    \end{minipage}
    \hfill
    \begin{minipage}{0.48\textwidth}
        \centering
        \includegraphics[width=0.7\linewidth]{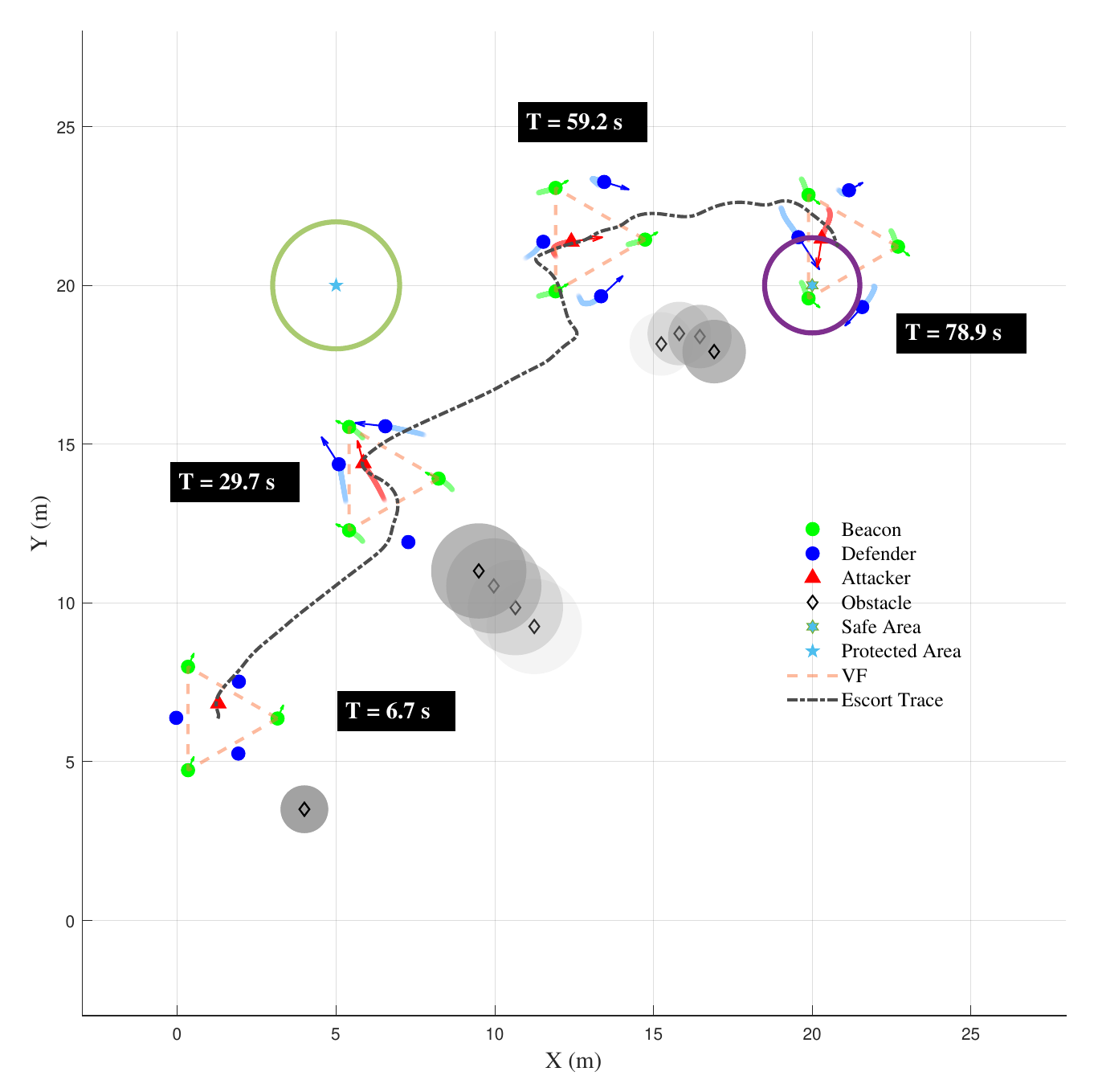}
        \subcaption{}
        \label{fig:stage2}
    \end{minipage}
    \caption{Snapshots of the players' trajectory in the capture stage and the escort stage. (a) Capture Stage; (b) Escort Stage}
    \label{fig:Defense Trace}
    \vspace{-2mm}
\end{figure}

\begin{figure*}[!htbp]
    \centering
    \begin{minipage}{0.24\textwidth}
        \centering
        \includegraphics[width=1\linewidth]{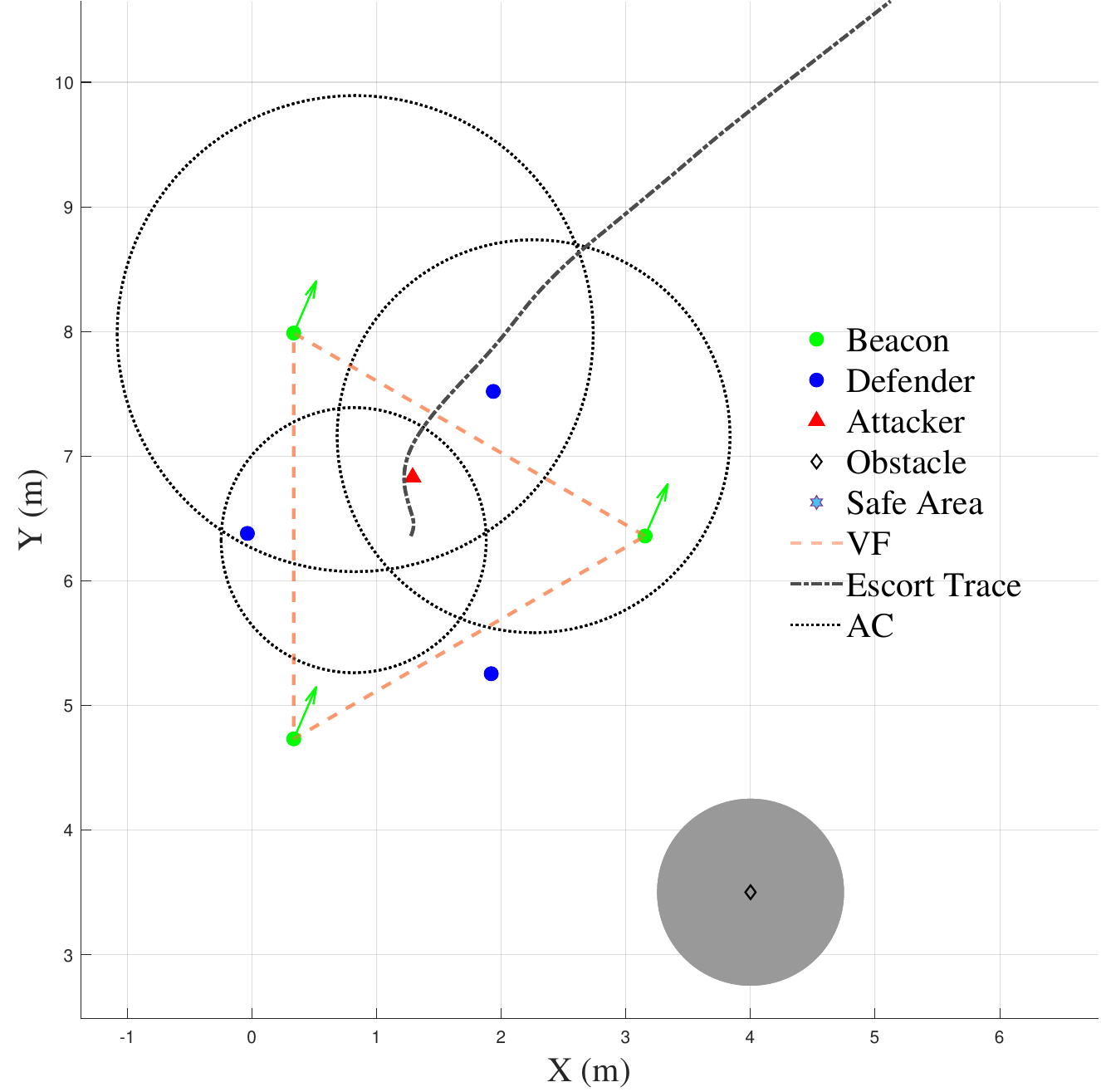}
        \subcaption{t=6.7s}
        \label{fig:T1}
    \end{minipage}\hspace{0.01\textwidth}%
    \begin{minipage}{0.24\textwidth}
        \centering
        \includegraphics[width=1\linewidth]{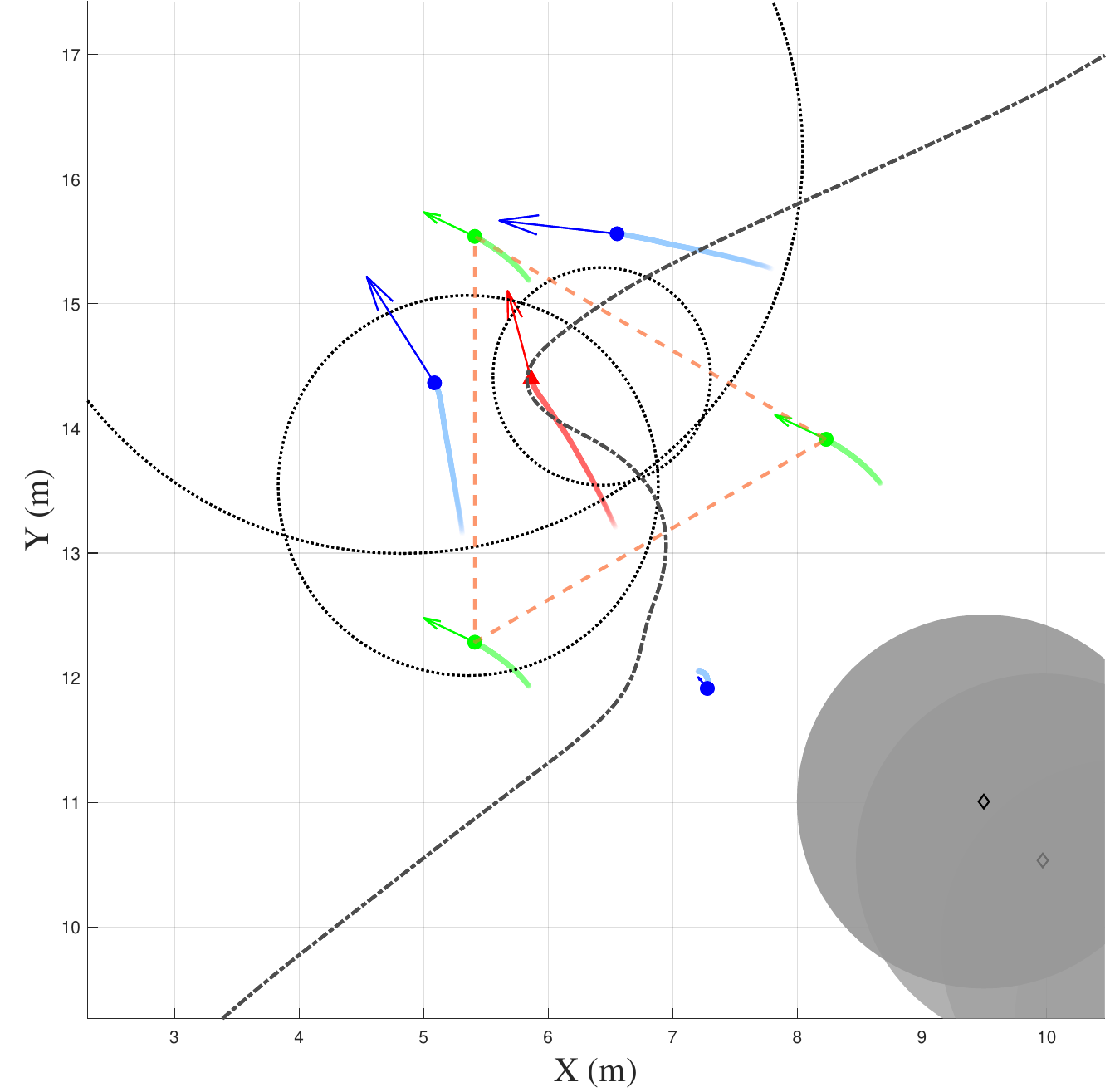}
        \subcaption{t=29.7s}
        \label{fig:T2}
    \end{minipage}\hspace{0.01\textwidth}%
    \begin{minipage}{0.24\textwidth}
        \centering
        \includegraphics[width=1\linewidth]{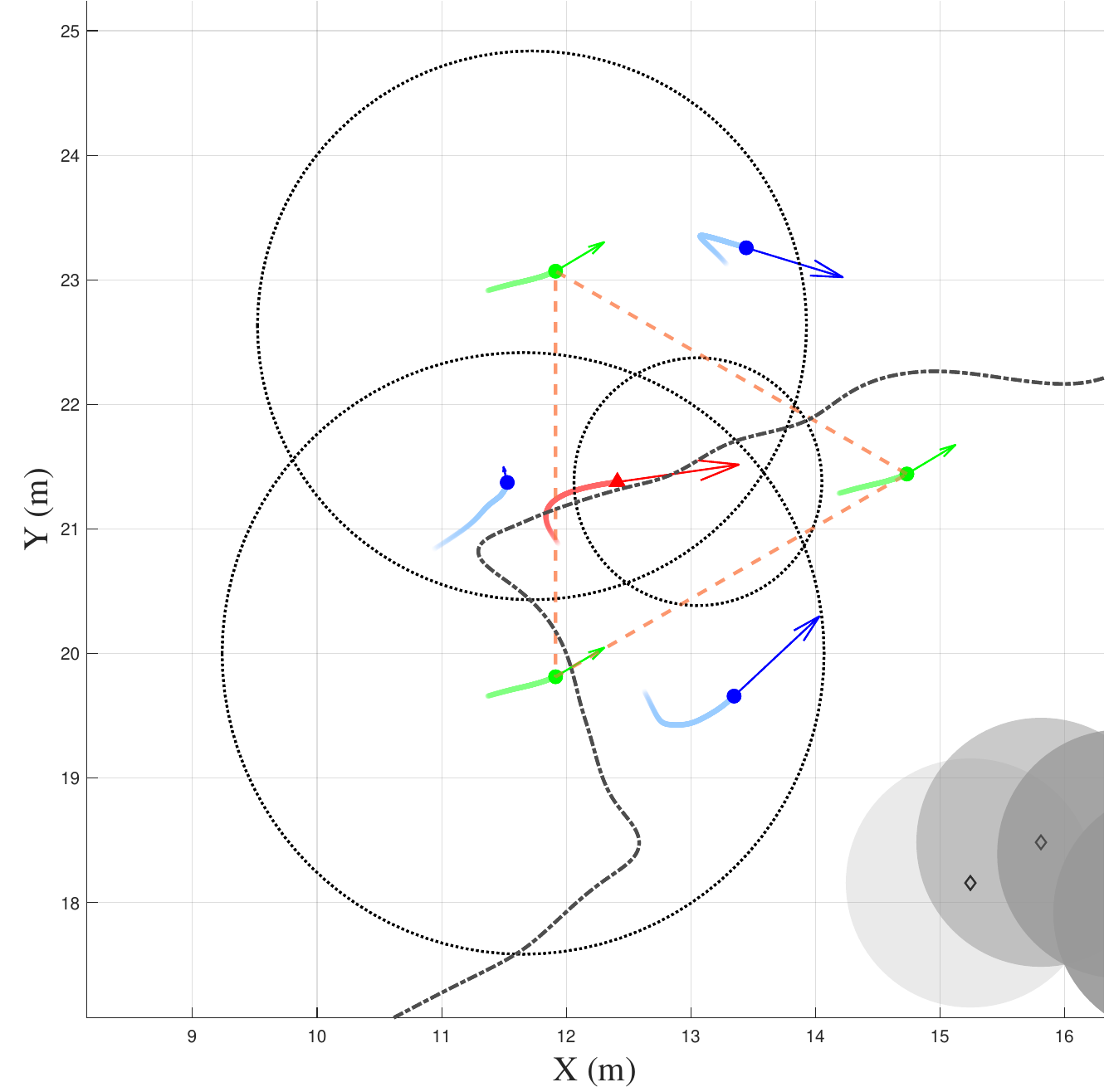}
        \subcaption{t=59.2s}
        \label{fig:T3}
    \end{minipage}\hspace{0.01\textwidth}%
    \begin{minipage}{0.24\textwidth}
        \centering
        \includegraphics[width=1\linewidth]{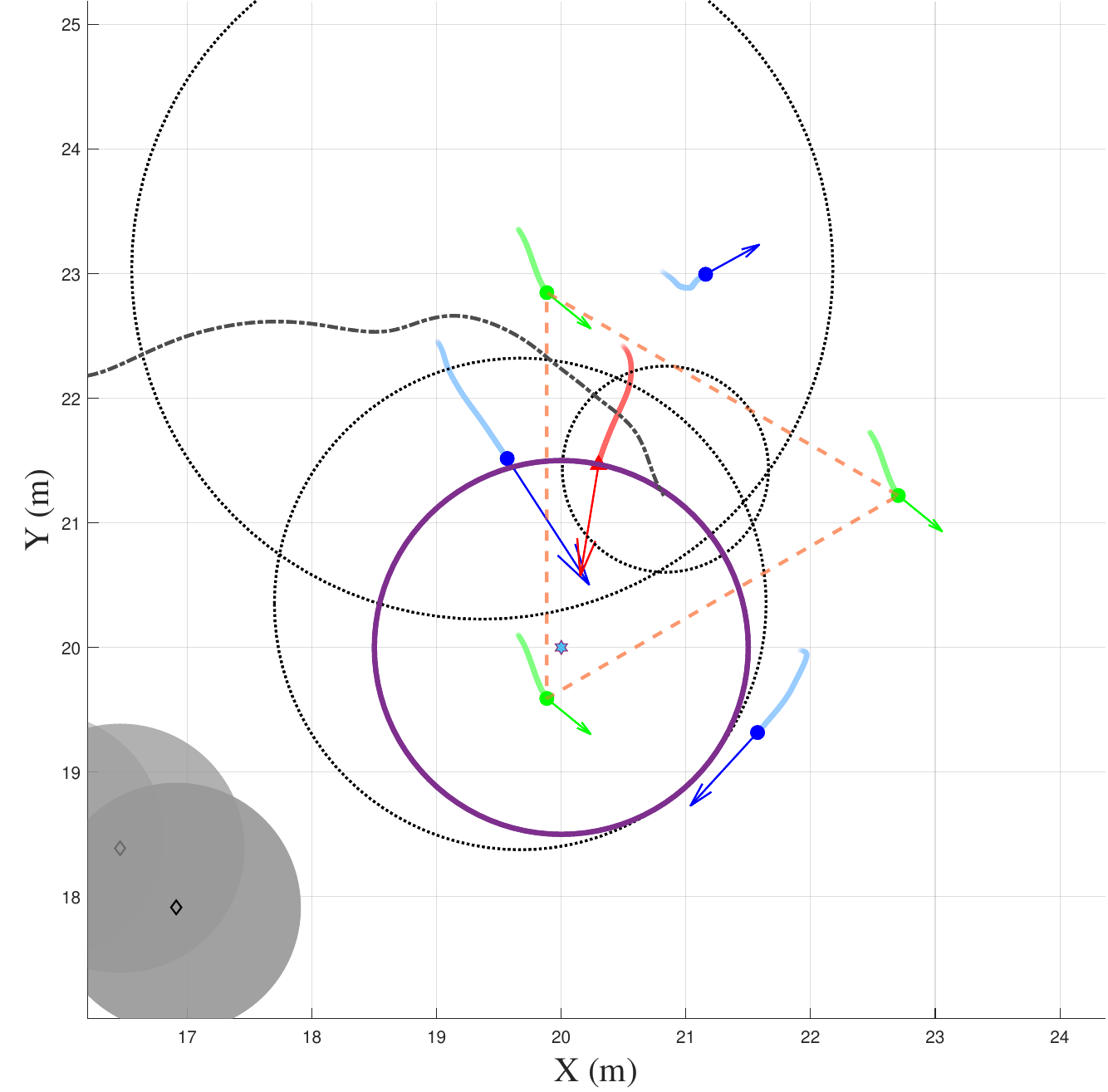}
        \subcaption{t=78.9s}
        \label{fig:T4}
    \end{minipage}
    \caption{Snapshots of reachable set visualization}
    \label{fig: Snapshots of reachable set visualization}
\end{figure*}

\begin{figure}[!htbp]
    \vspace{-2mm}
\centering
\includegraphics[width=1\linewidth]{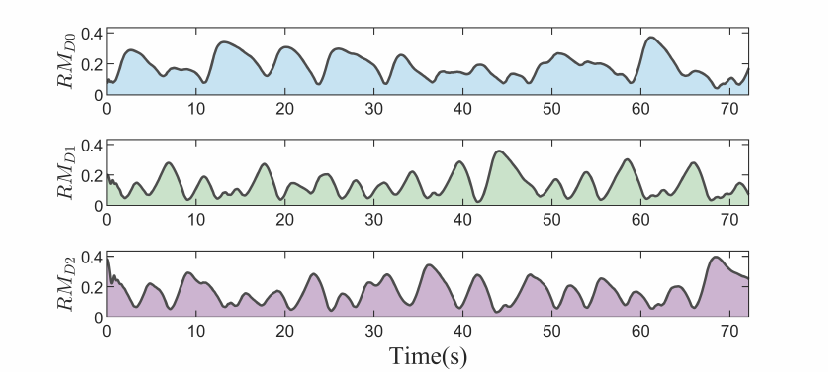}
\caption{The risk margin of game situation}
\label{fig: The risk margin of game situation}
\end{figure}

Fig. \ref{fig:Defense Trace} depicts the movement trajectory of the players throughout the entire guidance process. Note that the Times marked in the figure are respectively the elapsed times relative to the beginning moment of each stage. As shown in Fig. \ref{fig:stage1}, The defenders set out in an unstructured formation, and as the pursuit circle is updated, they gradually surround the attacker. Eventually, while forming the preset initial configuration $\mathcal{F}^{init}$ that satisfies (\ref{formation param calculation}) and the terminal speeds become $0$, the capture stage is over. At this time, the distance between the adjacent defenders is $\sqrt{3}\epsilon_D=2.27m>1.6m$, if the traditional sheep herding method based on fixed formations is adopted, the attackers are very likely to pass between the defenders, resulting in the failure of herding. Fig. \ref{fig:stage2} shows the escort process in a complex environment with static and dynamic obstacles, and the grey dotted lines represent the trajectory of the center point of the virtual fence. It can be found that under the action of the joint virtual force, the beacons can effectively guide the defender group away from the obstacles, which only relies on the local environmental information detected online. Further, each snapshot in Fig. \ref{fig: Snapshots of reachable set visualization} highlights multiple attack directions of the attacker and the corresponding interception responses of the defenders, for each $D_i-A(i \in I_d)$ pair, the interior of the apolloniz circle is denoted as $\mathcal{R}^i_{AC}$(Black short dotted line), then the reachable set of $A$ is $\mathcal{R}_A = \mathcal{R}^0_{AC} \cap \mathcal{R}^1_{AC} \cap \mathcal{R}^2_{AC}$, we observe $\mathcal{R}_A \cap \mathcal{E}_N = \varnothing$, i.e., the attacker cannot escape the virtual fence, the positive value result of the judgment function shown in Fig. \ref{fig: The risk margin of game situation} further proves this point. Fig. \ref{fig:T1} and Fig. \ref{fig:T4} are the initial and final moments of the escort stage. Fig. \ref{fig:T2} depicts the scenario where the attacker attempts to breach the virtual fence from a vertex of $\mathcal{E}_N$. The polygonal structure of the fence enables two defenders to collaboratively intercept the attacker under the game-layer control strategy. In contrast, Fig. \ref{fig:T3} shows the attacker attempting to breach along an edge of $\mathcal{E}_N$, Unlike the vertex scenario, a single defender is sufficient in this case to force the attacker to alter its movement direction, thereby redirecting it inward toward the virtual fence enclosure.

\begin{figure}[!htbp]
    \vspace{-2mm}
    \centering
    \includegraphics[width=1\linewidth]{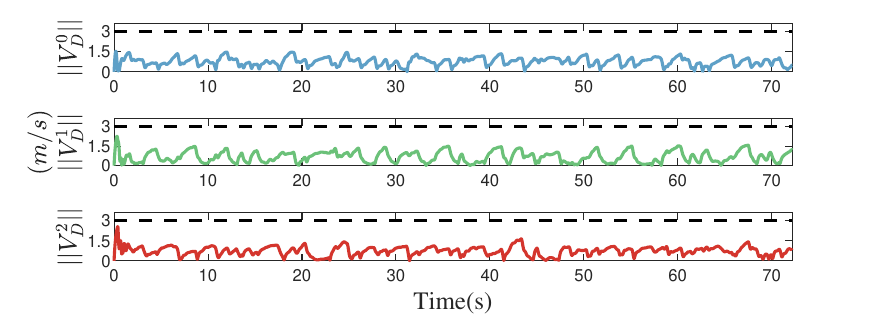}
    \caption{The euclidean norm of defenders' velocity}
    \label{fig: velocity}
\end{figure}

\begin{figure}[!htbp]
    \vspace{-2mm}
\centering
\includegraphics[width=1\linewidth]{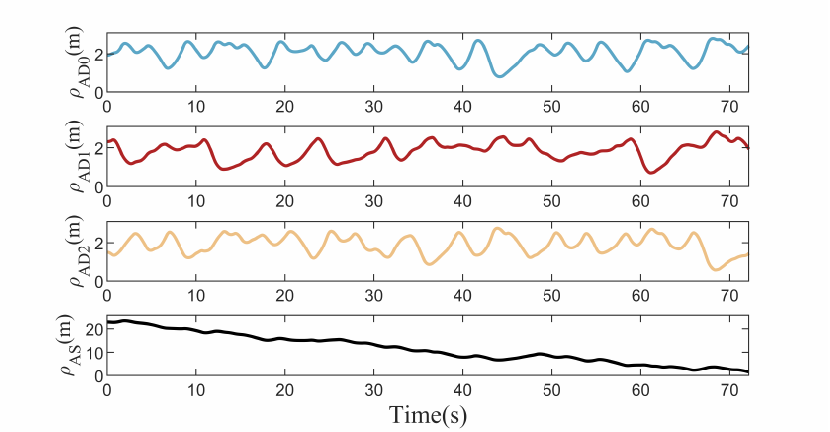}
\caption{The relative distance from defenders to attacker and attacker to safe area}
\label{fig: relative distance}
\end{figure}

\section{Conclusion}
This paper investigates the indirect guidance problem in complex dynamic environments. We decompose the entire process into capture stage and escort stage, and further break down the multi-player reach-avoid (RA) game during escort stage into pairwise sub-games on the virtual fence boundary. Based on reachability analysis, we develop defenders' capture and interception strategies, along with a virtual fence moving strategy, to ensure the safe migration of attackers with unknown strategies into predetermined regions. The proposed methodology is analytical, offers excellent computational efficiency, and requires minimal parameter tuning, making it suitable for real-world deployment. Future work will focus on extending indirect herding approaches to scenarios involving larger agent populations and higher-dimensional spaces.

\appendices
\section{Proof of Theorem 1}
In an unbounded space with a defense line $DL$ that separates $A$ and $D$ on opposite sides, according to Lemma 1, the reachable region $\mathcal{R}_A$ is an Apollonius circle determined by the relative positions and speed ratio of $A$ and $D$. When the Apollonius circle and $DL$ have no intersection points, the defender can always reach $DL$ before the attacker, implying that $A$ cannot cross $DL$ without being intercepted by $D$. Consider the critical scenario in which $\mathcal{R}_A$ and $DL$ are tangent at a single point $\boldsymbol{p^*}$, referred to as the Optimal Target Point (OTP) for $A$ \cite{yanReachAvoidGamesTwo2019}. The corresponding relative position condition yields the analytical expression for the barrier $\mathcal{B}_D$.\par
\begin{figure}[!htbp]
    \centering
    \includegraphics[width=0.7\linewidth]{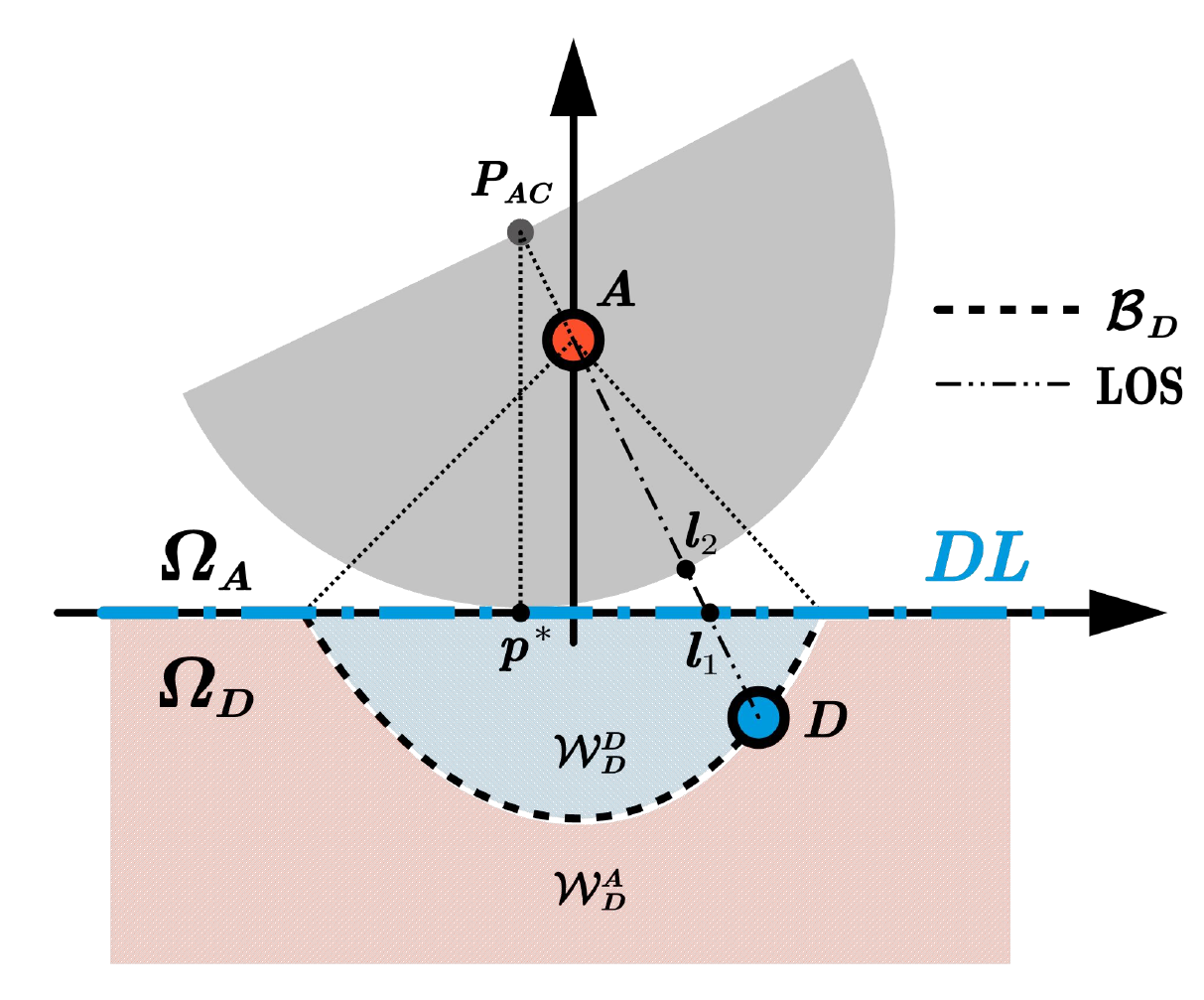}
    \caption{Barrier}
    \label{fig: barrier}
\end{figure}
Let $P_{AC}$ with radius $R_{AC}$ represent the center of the Apollonitz Circle determined by (\ref{Apollonius Circle}), the vertical distance from $P_{AC}$ to $DL$ is $l_p$, and it satisfies
\begin{equation}\label{lp}
\begin{aligned}
    \frac{l_p}{\sin{\varphi}} &= R_{AC} + \|\mathbf{x}_A-\boldsymbol{l}_1\|-\|\mathbf{x}_A-\boldsymbol{l}_2\|\\
        &= \frac{\alpha_V}{1-\alpha_V^2}\rho_{a,d}+\frac{l_a}{\sin{\varphi}}-\frac{\alpha_V}{1+\alpha_V}\rho_{a,d}\\
        &= \frac{\alpha_V^2}{1-\alpha_V^2}\rho_{a,d}+\frac{l_a}{\sin{\varphi}}
\end{aligned}
\end{equation}
where $\rho_{a,d} = \|\mathbf{x}_A-\mathbf{x}_D\|=(l_d+l_a)/\sin{\varphi}$ denotes the relative distance between $A$ and $D$. Substituting $\rho_{a,d}$ into (\ref{lp}), $l_p$ can be formulated as
\begin{equation}
    l_p = \frac{\alpha_V^2}{1-\alpha_V^2}l_d+\frac{1}{1-\alpha_V^2}l_a
\end{equation}
It's clearly that if $l_p-R_{AC}=0$, $\mathcal{R}_A \cap DL = \boldsymbol{p^*}$, if $l_p-R_{AC}>0$, $\mathcal{R}_A \cap DL = \varnothing$. Let $RM$ be the value of $l_p-R_{AC}$, which represents the state function of the game, then we have
\begin{equation}
\begin{aligned}
    RM &= \frac{\alpha_V^2}{1-\alpha_V^2}l_d+\frac{1}{1-\alpha_V^2}l_a - \frac{\alpha_V}{1-\alpha_V^2}\frac{l_a+l_d}{\sin{\varphi}}\\
                & = \frac{1}{1-\alpha_V^2}[l_a(1-\frac{\alpha_V}{\sin{\varphi}})-l_d(\frac{\alpha_V}{\sin{\varphi}}-\alpha_V^2)]\\
                &=\frac{1}{1-\alpha_V^2}J(l_a,l_d,\varphi)
\end{aligned}
\end{equation}

Because $1/(1-\alpha_V^2)$ is a positive constant, the symbol for $RM$ is the same as that for $J$. Fix the position of $A$, $J \equiv 0$ determines a one-dimensional curve in the defender's movement space $\Omega_{D}$, i.e., the defender's barrier $\mathcal{B}_D$ shown in Fig. \ref{fig: barrier}.

\section{Proof of Theorem 2}
Given the symmetry of polygons, we can analyze the configuration parameters from just one sector of PC (dark areas in the figure). For an $N$-sided polygon, the spread angle corresponding to the sector is $\lambda = 2\pi/N$, and the boundary and interior of the sector are locations where attackers may be present. Furthermore, to avoid singularities, we also appropriately expand the PC to obtain the Extended PC (E-PC) with a zoom factor $K_p > 1$, the corresponding sector is marked as $S^{EPC}_i$.\par
\begin{figure}[!htbp]
\centering
\includegraphics[width=0.7\linewidth]{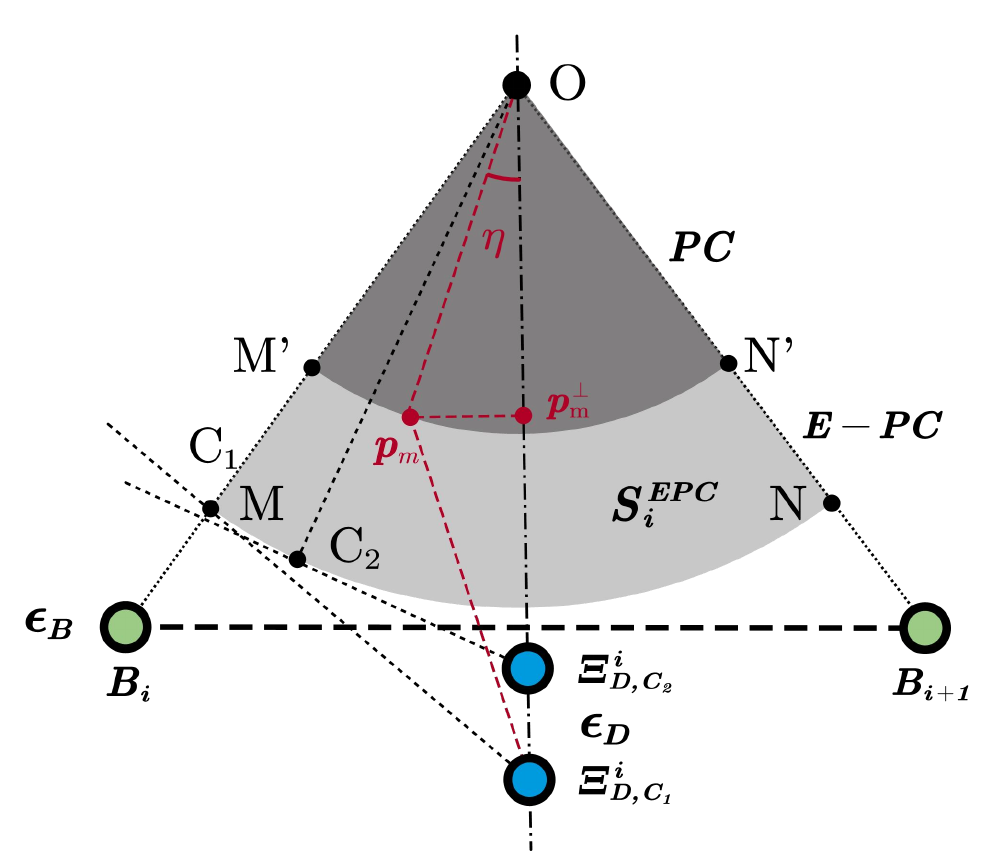}
\caption{Schematic diagram of geometric relationships within sectors}
\label{fig: Therom2}
\end{figure}
To minimise the initial configuration $\mathcal{F}^{init}$, for all points in E-PC, we first need to find the minimum $\epsilon_D$ that satisfies \textbf{Condition 1} in (\ref{sufficient equations}), i.e.
\begin{equation}\label{AppendixB.1}
    \begin{aligned}
        & \mathop{\text{minimize}}\limits_{\forall \mathbf{x}_A \in S^{EPC}_i}\; \epsilon_D \\
        & \text{s.t.} \quad  \Psi \leqslant \varphi \leqslant\pi-\Psi
    \end{aligned}
\end{equation}
where $\Psi = \sin^{-1}{\alpha_V}$ represents the lower bound of the LOS angle $\varphi$ from $\Xi^i_D$ to $\mathbf{x}_A$. Construct a tangent line $TL$ rolling on the boundary $\partial S^{EPC}_i$, and the angle between $TL$ and $DL$ is $\Psi$. The intersection point $\Xi^i_D$ of $TL$ and the angle bisector is the point that satisfies (\ref{AppendixB.1}), and the tangent point $C$ between $TL$ and $\partial S^{EPC}_i$ is referred to as Critical LOS Point ($CLP$). From Fig. \ref{fig: Therom2}, it is easy to observe that there are two possible cases: $\Psi$ is less than or equal to $\lambda/2$, or $\Psi$ is greater than $\lambda/2$, corresponding to different $CLP$. For $\Psi > \lambda/2$, $CLP$ is the fixed boundary point $C_1$ ($M$), for $\Psi \leqslant \lambda/2$, $CLP$ is a tangent point $C_2$ located in the arc segment $\overset{\frown}{MN}$. In the first case, two interior angles $\angle C_1O\Xi^i_{D,C_1}$ and $\angle C_1\Xi^i_{D,C_1}O$ are $\lambda/2$ and ($\pi/2-\Psi$) respectively, so $\angle OC_1\Xi^i_{D,C_1} = (\pi/2+\Psi-\lambda/2)$, the radius of E-PC ($\overline{OC_1}$) is $K_p\epsilon_P$, according to the sine theorem, $\overline{O\Xi^i_{D,C_1}}=\epsilon_D$ can be expressed as
\begin{equation}\label{AppendixB.2}
\begin{aligned}
    \overline{O\Xi^i_{D,C_1}} & = \frac{\overline{OC_1}}{\sin(\angle C_1\Xi^i_{D,C_1}O)}\sin(\angle OC_1\Xi^i_{D,C_1})\\
    & = K_p\epsilon_P[\sin(\frac{\lambda}{2})\tan\Psi+\cos(\frac{\lambda}{2})]
\end{aligned}
\end{equation}
In the second case, $\triangle OC_2\Xi^i_{D,C_2}$ is a right-angle triangle with $\angle OC_2\Xi^i_{D,C_2}$ being $\pi/2$, where the right-angle side $\overline{OC_2}$ is $K_p\epsilon_P$ and its opposite angle $\angle C_2\Xi^i_{D,C_2}O$ is $(\pi/2-\Psi)$, then $\overline{O\Xi^i_{D,C_2}}=\epsilon_D$ can be expressed as
\begin{equation}\label{AppendixB.3}
    \overline{O\Xi^i_{D,C_2}} = \frac{\overline{OC_2}}{\sin(\angle C_2\Xi^i_{D,C_2}O)} = \frac{K_p\epsilon_P}{\cos{\Psi}}
\end{equation}
Note that $\sin{\Psi} = \alpha_V$ and $\cos{\Psi} = \sqrt{1-\alpha_V^2}$, by substituting them into (\ref{AppendixB.2}) and  (\ref{AppendixB.3}) yields (\ref{formation param calculation.1}).\par
Then fix $\epsilon_D$ as a preset constant for determining the minimum size of the Virtual Fence, i.e., $\epsilon_B$, to satisfy \textbf{Condition 2} in (\ref{sufficient equations}). Consider a moving point $\boldsymbol{p}_m$ on $\overset{\frown}{M'N'}$ as the possible critical position of the attacker, and the angle between $O\boldsymbol{p}_m$ and the angle bisector is $\eta$. The distance between $\boldsymbol{p}_m$ and $\Xi^i_D$ is denoted as $\rho_{p,\Xi}$, and the vertical distance from them to $DL_i$ are $l_{p}$ and $l_{\Xi}$, respectively. Our goal can be expressed as
\begin{equation}\label{AppendixB.4}
    \begin{aligned}
       & \mathop{\text{minimize}}\limits_{\forall \eta \in [0,\lambda/2]} \quad \epsilon_B\\
       & \text{s.t.} \quad J(l_p,l_{\Xi},\varphi_{p,_\Xi}) \geqslant 0
    \end{aligned}
\end{equation}
Where $\varphi_{p,\Xi}$ denotes the LOS angle between $\boldsymbol{p_m}$ and $\Xi^i_D$. Note that $\epsilon_D$ has been fixed, so (\ref{AppendixB.4}) can be equivalently converted to finding the minimum value of $l_{\Xi}$  that satisfies $J=0$ when $\eta$ varies within the range $[0,\lambda/2]$. According to geometric relationships and the cosine theorem, we have
\begin{equation}\label{AppendixB.5}
    \begin{aligned}
        & \rho_{p,\Xi}(\eta) = \sqrt{\epsilon_P^2+\epsilon_D^2-2\epsilon_P\epsilon_D\cos{\eta}} \\
        & \rho_{p,\Xi}\sin{\varphi_{p,\Xi}} = \epsilon_D-\epsilon_P\cos{\eta}
    \end{aligned}
\end{equation}
substitute (\ref{AppendixB.5}) into (\ref{Barrier Fcn}), we have
\begin{equation}\label{AppendixB.6}
    \begin{aligned}
        l_{\Xi}(\eta) = & \frac{1}{1-\alpha_V^2}(\epsilon_D-\epsilon_P\cos{\eta} \\
        & -\alpha_V\sqrt{\epsilon_P^2+\epsilon_D^2-2\epsilon_P\epsilon_D\cos{\eta}})
    \end{aligned}
\end{equation}
Differentiating (\ref{AppendixB.6}) with respect to $\eta$, we further obtain
\begin{equation}\label{AppendixB.7}
    \begin{aligned}
        \frac{dl_{\Xi}}{d\eta} = & \frac{1}{1-\alpha_V^2}(\epsilon_P\sin{\eta} \\
        & - \alpha_V\frac{\epsilon_P\epsilon_D\sin{\eta}}{\sqrt{\epsilon_P^2+\epsilon_D^2-2\epsilon_P\epsilon_D\cos{\eta}}})
    \end{aligned}
\end{equation}
$l_{\Xi}(\eta)$ takes the extreme value when $dl_{\Xi}/d\eta = 0$ or at the boundary $\eta_1 = 0$, $\eta_2 = \lambda/2$. For $dl_{\Xi}/d\eta = 0$, (\ref{AppendixB.7}) has two solutions
\begin{subequations}
    \begin{align}
        & (1)\sin{\eta} = 0 \label{Solution1}\\
        & (2)\cos{\eta}=\frac{[\epsilon_P^2+(1-\alpha_V^2)\epsilon_D^2]}{2\epsilon_P\epsilon_D} = \chi\label{Solution2}
    \end{align}
\end{subequations}
Since $0 \leqslant \eta \leqslant \lambda/2 < \pi/2$, trigonometric functions are monotonic, so if there exist real solution, i.e. $\chi \leqslant 1$, both (\ref{Solution1}) and (\ref{Solution2}) have a single solution, $\eta_1 = 0$ and $\eta^* = \cos^{-1}\chi$, respectively. The minimum value of $l_{\Xi}$ can be expressed as
\begin{equation}\label{Minimum l_Xi}
    l_{\Xi,\min} = \min[\, l_{\Xi}(0),l_{\Xi}(\eta^*),l_{\Xi}(\lambda/2) \,]
\end{equation}
for $\chi > 1$ or $\cos^{-1}\chi > \lambda/2$, (\ref{Minimum l_Xi}) is rewritten as
\begin{equation}
    l_{\Xi,\min} = \min[ \, l_{\Xi}(0),l_{\Xi}(\lambda/2) \, ]
\end{equation}

\section{Proof of Theorem 3}
Consider the R-A sub-games described by (\ref{sufficient equations}) in the local coordinate system fixedly connected to the virtual fence, both the defenders and the attackers superimposed the relative velocity $\mathbf{v}_{Fc}$. Let the speed of the defender in this coordinate system be represented as $\widetilde{\mathbf{v}}_D$ and the speed of the attacker as $\widetilde{\mathbf{v}}_A$, we have $\widetilde{\mathbf{v}}_D = \mathbf{v}_D - \mathbf{v}_{Fc}$ and $\widetilde{\mathbf{v}}_A = \mathbf{v}_A - \mathbf{v}_{Fc}$. Note that we have the design speed ratio $\hat{\alpha_V} \in (\alpha_V,1)$, it means that when the local actual velocity ratio $\left. \frac{\|\widetilde{\mathbf{v}}_A\|}{\|\widetilde{\mathbf{v}}_D\|}\right|_{\|\mathbf{v}_A\|=\overline{V}_A,\|\mathbf{v}_D\|=\overline{V}_D}$ in the sub-game is less than $\hat{\alpha_V}$, the winning-guaranteed strategy we have constructed is still valid. Obviously, the maximum speed of the virtual fence must be between that of the defender and the attacker, then the modulus of velocity satisfies the following relationship
\begin{equation}\label{AppendixC.1}
    \begin{aligned}
        & \overline{V}_A-\|\mathbf{v}_{Fc}\| \leqslant \|\widetilde{\mathbf{v}}_A\| \leqslant  \overline{V}_A+\|\mathbf{v}_{Fc}\|\\
        & \overline{V}_D-\|\mathbf{v}_{Fc}\| \leqslant \|\widetilde{\mathbf{v}}_D\| \leqslant  \overline{V}_D+\|\mathbf{v}_{Fc}\|
    \end{aligned}
\end{equation}
The upper bound of the local actual velocity ratio can be obtained from (\ref{AppendixC.1})
\begin{equation}\label{AppendixC.2}
    \frac{\|\widetilde{\mathbf{v}}_A\|}{\|\widetilde{\mathbf{v}}_D\|} \leqslant \frac{\overline{V}_A+\|\mathbf{v}_{Fc}\|}{\overline{V}_D-\|\mathbf{v}_{Fc}\|}
\end{equation}
then we can obtain the constraint of the beacon speed
\begin{equation}
    \begin{aligned}
        & \frac{\overline{V}_A+\|\mathbf{v}_{Fc}\|}{\overline{V}_D-\|\mathbf{v}_{Fc}\|} \leqslant \hat{\alpha_V} \Leftrightarrow \\
        & \|\mathbf{v}_{Fc}\| \leqslant \overline{V}_B \triangleq \min(\frac{\hat{\alpha_V}\overline{V}_D-\overline{V}_A}{1+\hat{\alpha_V}},\overline{V}_A)
    \end{aligned}
\end{equation}

\ifCLASSOPTIONcaptionsoff
  \newpage
\fi

\bibliographystyle{IEEEtran}
\bibliography{MyRef}

\end{document}